\documentclass[11pt]{article}

\usepackage[preprint]{acl}

\usepackage{times}
\usepackage{latexsym}
\usepackage{placeins}
\definecolor{trdeb}{HTML}{1F77B4}
\definecolor{trmono}{HTML}{FF7F0E}
\definecolor{trrest}{HTML}{2CA02C}
\definecolor{msol}{HTML}{0072B2}\definecolor{mluna}{HTML}{E69F00}\definecolor{mmini}{HTML}{009E73}
\definecolor{mds}{HTML}{CC79A7}\definecolor{mhaiku}{HTML}{D55E00}\definecolor{mqwen}{HTML}{56B4E9}

\usepackage[T1]{fontenc}
\usepackage[utf8]{inputenc}

\usepackage{microtype}

\usepackage{inconsolata}

\usepackage{graphicx}

\usepackage{booktabs}
\usepackage{multirow}
\usepackage{xcolor}
\usepackage{amsmath}
\usepackage{float}
\usepackage{tabularx}
\usepackage{array}
\usepackage{tikz}
\definecolor{popblue}{RGB}{0,114,178}
\definecolor{perorange}{RGB}{213,94,0}

\title{From Simulated Citizens to Simulated Deliberation:\\ Challenges in Representation and Interaction}

\author{
Chaemin Jang$^{1}$, Junsik Min$^{1}$, Jaewoo Choi$^{1}$, Donggyu Lee$^{1}$, \\ Haiin Lee$^{1}$, Yoonseo Choi$^{1}$,  
Junyoung Park$^{2}$, Namhee Kim$^{2}$, \\ Hyunwoo Kim$^{1}$, Juho Kim$^{1}$\footnotemark[2], Nuri Kim$^{1}$\footnotemark[2], Jihee Kim$^{1}$\footnotemark[2] \\
$^{1}$KAIST \quad $^{2}$Seoul National University \\
\texttt{jchaemin@kaist.ac.kr}
}

\begin{document}
\maketitle
\renewcommand{\thefootnote}{\fnsymbol{footnote}}
\footnotetext[2]{Co-corresponding authors.}
\renewcommand{\thefootnote}{\arabic{footnote}}

\begin{abstract}

Can LLM agents simulate how a real public would deliberate?
For such simulations to be meaningful, agents must reflect the population's opinions, and their views must be shaped by what others say during discussion. We evaluate these requirements across six LLMs using national surveys, controlled deliberation experiments, and a real deliberative poll. Comparing census-grounded Korean personas against national surveys, we find that agents often produce overly concentrated opinions and fail to reliably reproduce demographic differences.
During simulated deliberation, agents exchange arguments, respond to one another, and frequently change positions.
Yet similar changes occur even when agents cannot see one another's contributions, while simply restating their starting positions sharply reduces movement.
We further test whether these limitations persist in a real-world setting, using agents constructed from actual respondents in the 2017 Shin-Kori deliberative poll.
Even with real respondent profiles, agents fail to reproduce the panel's initial opinion distribution or how its opinions change across survey waves.
Our findings show that generating deliberative discussions, representing public opinion, and reproducing interaction-driven opinion change are distinct capabilities.
These results suggest a different potential role for LLM deliberation: surfacing arguments and perspectives, rather than simulating a real public.

% The apparent quality of a simulated discussion is therefore not sufficient evidence that it captures how a real public would deliberate.

% The ability to simulate deliberative discourse should not be mistaken for the ability to simulate a deliberating public.
% These findings point to a different potential role for LLM deliberation: surfacing arguments and perspectives around contested issues
\end{abstract}

\section{Introduction}
Public deliberation does more than aggregate what citizens already think. By exposing participants to balanced information, competing arguments, and perspectives different from their own, deliberation is thought to produce more considered judgments at the individual level and better collective decisions at the societal level \citep{cohen1989,dryzek2000,fishkin2009,gutmannthompson1996,habermas1989,mansbridge2010}. The process itself is also revelatory. It surfaces  arguments and evidence behind disagreement across groups, giving policymakers a map of where contention lies and citizens a structured view of positions other than their own \citep{fung2003,fishkin2009}. 
However, realizing this ideal at scale is costly and the time and resources it requires limit how often and on what scale such processes can be run \citep{fishkin2009}.

Large language models offer a tempting substitute. One can condition LLM agents on realistic citizen personas, let them debate a policy agenda, and inspect the result \citep{park2023,argyle2023}. The promise is an efficient preview of the deliberative process. For example, policymakers could simulate the process to see which arguments emerge, which groups might align, where disagreement persists, and how views evolve after exposure to competing reasons. Recent work has begun to explore adjacent uses, including AI mediation that
helps human groups find common ground \citep{tessler2024} 
and language-model support for scalable deliberation platforms
\citep{small2023polis}. Such a simulation would be useful not merely if it predicts which side wins, but if it captures which social perspectives enter the discussion, which arguments are considered, and how views change through the exchange. The question is, then, what must be validated before LLM-based deliberation can support these uses. 

What would it mean to simulate deliberation well? A simulated deliberation can be judged in two ways: normatively, by whether it approximates a balanced, inclusive, and reason-responsive public sphere \citep{fishkin2005}, or descriptively, by whether it resembles how people actually deliberate closely enough to transfer to real settings. 
For a simulation intended to represent how a target population would deliberate, two basic conditions are necessary:

\begin{itemize}
    \item[(A1)] 
    \textbf{Opinion representation.}
Persona agents should reproduce how opinions are distributed across demographic groups in the population they represent.
  \item[(A2)] \textbf{Interaction.} Agents should respond to one another's contributions, so that the exchange shapes what they consider and ultimately conclude.
\end{itemize}
If (A1) fails, a demographically representative simulation can fail to represent the population it claims to stand for because it gets the relationship between demographics and opinions wrong.
If (A2) fails, apparent opinion change may arise without being driven by peer exchange, so stance movement alone does not establish interaction. We test both conditions in a Korean policy setting. If a simulation is to stand in for a public, we can further ask whether its agents change as a deliberating public did. We test this against the 2017 Shin-Kori deliberative poll on two nuclear reactors \citep{shingori2017}, whose committee released the first-wave survey the panel was drawn from: we build agents from real respondents' first-wave records, run the same protocol, and compare each stage with the real panel. 

We find substantial evidence against (A1): persona agents do not reliably reproduce the population opinion patterns they are meant to represent. Their responses are often much more concentrated than the human survey, and the demographic groups that are more or less supportive frequently differ from those in the survey. Testing (A2) reveals a different problem. The agents do produce reciprocal exchanges and often revise their stated positions, but much of the observed movement can arise without peer exchange: sealed-monologue controls, in which agents never see one another's turns, produce nearly the same average final room composition in five of six models. Against the real poll, agents built from real respondents' records miss the first wave and discount arguments attributed to peers. These results show that the visible features of deliberation, population alignment, and peer-interaction-driven stance change are distinct properties of LLM deliberation.
% A single regularity underlies these results: agents hold to their own stated positions and discount what peers say. These results show that the visible features of deliberation, population alignment, and peer-interaction-driven stance change can come apart.

Our study makes four contributions to the evaluation of LLM-based deliberation.
\begin{itemize}
\item \textbf{We provide a survey-grounded evaluation of persona agents.} We benchmark them against national surveys and test whether the mismatch persists across persona specifications, response formats, populations, and models (Section~\ref{sec:respondents}).
\item \textbf{We separate apparent deliberative change from the effect of peer interaction.} We test whether agents' stance changes depend on hearing and responding to other agents or can arise without peer exchange (Section~\ref{sec:deliberation}).
\item \textbf{We test simulated deliberation against a real deliberative poll.} Building agents from the 2017 Shin-Kori deliberative poll, we run the same survey and deliberation protocol and compare each stage with the panel's survey waves (Section~\ref{sec:poll}).
\item \textbf{We distinguish population simulation from argument surfacing.} We show that agents can produce reciprocal, reasoned, and varied arguments even when their opinion distributions do not reliably represent the population. This motivates treating argument surfacing and population simulation as distinct uses with different validation requirements (Sections~\ref{sec:deliberation}, \ref{sec:discussion}).
\end{itemize}

\begin{figure*}[t]
\vspace{-3mm}
\centering
\includegraphics[width=\linewidth]{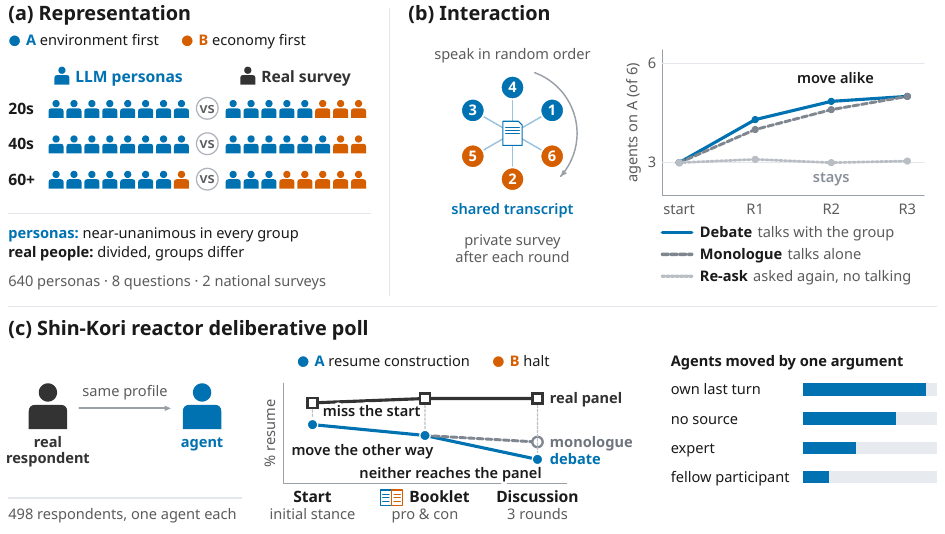}
\vspace{-3mm}
\caption{Experimental setup. \textbf{(a)} A mini-public of 640 Korean personas, drawn from Nemotron-Personas-Korea and jointly balanced on sex, age, education, and region, answers eight two-position policy questions taken from two national surveys at temperature~0 (Section~\ref{sec:setup}). \textbf{(b)} Six personas then deliberate for three rounds over a cumulative shared transcript: in each round every agent reads the full log, speaks once, and appends its turn. \textbf{(c)} Section~\ref{sec:poll} applies the same protocol to agents built from a real deliberative poll.}
\label{fig:overview}
\end{figure*}
\section{Related work}

\textbf{LLM agents as simulated publics.} LLM agents with personas and memory produce believable social behavior \citep{park2023}, and conditioning on demographic profiles can reproduce some aggregate survey distributions \citep{argyle2023}. Later work is more cautious: LLM
substitutes can misportray and flatten identity groups \citep{wang2025}. Applied systems include AI mediation of
human discussion \citep{tessler2024} and multi-agent deliberation built on ANES-based national personas \citep{ashkinaze2025}. We ask whether the
debate pipeline itself, applied to a census-grounded non-American population, meets the assumptions these uses require.

\textbf{Do persona LLMs represent a population?} On US survey data,
language-model opinions misalign with those of demographic groups
\citep{santurkar2023}, and survey-derived alignment is confounded by
response artifacts, format sensitivity, and instability across question
variants \citep{dominguez2024,moore2024}. We extend this line to
census-grounded Korean personas benchmarked group by group, with controls
over persona content, response instrument, language, and model scale.

\textbf{Biases in simulated debates.} 
\citet{taubenfeld2024} found that simulated partisan debaters drift toward
the base model's stance regardless of assigned persona, and
\citet{chuang2024} report related persona fragility in opinion-dynamics
simulations. A separate line applies multi-agent debate to tasks with
verifiable answers \citep{du2024,liang2024,chan2024}, where agreement is
evidence of correctness; that framing does not transfer to deliberation,
where no key exists.

\textbf{Measuring deliberation quality.} Deliberative polling evaluates
discussion against an ideal of balanced, reasoned exchange
\citep{fishkin2005}, the Discourse Quality Index codes each contribution for justification, respect, and common-good orientation \citep{steenbergen2003}, to which we add reciprocity, whether a turn responds to others \citep{graham2003,stromergalley2007}, and argument repertoire counts the distinct
reasons a participant gives for each side \citep{cappella2002}. We adapt
the last three as language-model-judge measures of whether the resulting
deliberation is good by established standards.
\section{Setup}
\label{sec:setup}
\label{sec:issues}\label{sec:personasetup}\label{sec:models}

\begin{table}[t]
\centering\small
\setlength{\tabcolsep}{3pt}
\begin{tabular}{@{}>{\raggedright\arraybackslash}p{1.55cm}>{\raggedright\arraybackslash}p{5.6cm}@{}}
\toprule
\textbf{Questions} & 8 two-position policy questions: 
4 environmental (KEI 2025, $n{=}3{,}008$; by sex, age, education, region) and 4 low-birthrate (PCASPP, $n{=}2{,}800$; by sex, age, marital status) \\
\addlinespace
\textbf{Personas} & 640 Nemotron-Personas-Korea profiles with a free-text narrative, 4 per sex $\times$ age $\times$ education $\times$ region cell; variants: demographics only, and 256 U.S. personas against Pew \\
\addlinespace
\textbf{Models} & GPT-6-sol (primary), GPT-6-luna, GPT-4.1-mini, DeepSeek-V4.1, Claude-Haiku-5.5, Qwen-3.8-flash; reasoning off; discourse judges Claude-Opus-5.5 and Gemini-3.1-pro \\
\addlinespace
\textbf{Survey} (\S\ref{sec:respondents}) & private choice between the two positions as plain statements, order randomized, temperature 0 \\
\addlinespace
\textbf{Debate} (\S\ref{sec:deliberation}) & 6 agents, 3 rounds, full transcript, temperature 1 \\
\addlinespace
\textbf{Real poll} (\S\ref{sec:poll}) & 2017 Shin-Kori deliberative poll: agents from 498 real first-wave records; compared with waves 1, 3, and 4 \\
\bottomrule
\end{tabular}
\caption{Study design.}
\label{tab:setup}
\end{table}

Table~\ref{tab:setup} summarizes the study.\footnote{Code, data, and all per-run outputs can be seen in https://github.com/jchaemin/simulated-deliberation.} The eight questions come from two Korean national surveys that publish benchmarks by demographic group; where an item has more than two options, two contrasting ones become Position~A and Position~B (wordings in Appendix~\ref{app:instruments}; Appendix~\ref{app:origformat} uses the original questionnaire). The persona pool is balanced rather than population-weighted, and we compare it with the survey within demographic groups; reweighting to the survey's composition barely changes the results (Appendix~\ref{app:robust}).

\section{Personas as survey respondents}
\label{sec:respondents}

If simulated deliberation is meant to represent a target population, its persona agents should first reproduce the population's baseline opinions. We test this before deliberation.

\subsection{Survey setup}
\label{sec:instrument}

Each persona chooses between the two positions of each question, presented as plain natural-language statements; we call the share choosing Position~A the \emph{A-share}. We survey all 640 personas of the Korean pool on the eight questions with GPT-6-sol at temperature~0; other models are in Appendix~\ref{app:robust}. To limit known presentation biases, including sensitivity to answer labels, option order, and scale framing \citep{dominguez2024,tjuatja2024,rottger2024}, the positions carry no labels or scales and appear in random order, and there is no neutral option, since the analysis concerns the relative choice between the two positions; Appendix~\ref{app:origformat} repeats it with the original survey formats.

We call a question \emph{divisive} if neither position receives more than 65\% of persona responses, and \emph{saturated} otherwise. Only environmental means is divisive (stronger penalties and law enforcement vs.\ voluntary efforts by citizens and firms first; A-share 43\%); the label describes the personas under this instrument, not the human benchmark.

\subsection{Persona responses do not reproduce population opinion patterns}
\label{sec:validity}

The persona responses fail to reproduce the survey's demographic patterns of opinion. Table~\ref{tab:bygroup} (Appendix~\ref{app:bygroup}) compares these pre-deliberation A-shares with the corresponding human survey benchmarks by demographic group. 
The failure holds in every group: the mean absolute gap across group cells is 29 percentage points, and every demographic group’s mean gap falls between 21 and 40 points. The answers are also extreme: on six of the eight questions, at least one group is near 0 or 100 while the corresponding human shares are far less concentrated. The degree of concentration varies substantially across questions, with some producing near-unanimous persona responses and others remaining relatively divided.

Under the original questionnaire, two personas from the same sex $\times$ age $\times$ education $\times$ region cell pick the identical pair of options 46 to 47\% of the time, against 4 to 8\% for two respondents from the same cell (Appendix~\ref{app:origformat}), so the personas also erase the variation among individuals within a group. 

Finally, the direction of demographic gaps is no better than chance: when the survey shows one group favoring a position more than another, the personas match that direction in only 49 of 110 comparisons and reverse it in 51, the remaining 10 being ties, and where the human difference exceeds 5 points they reverse it more often than they match it (19 against 14 of 39).
%The personas do not track the survey's demographic structure. 
%On average, a group's persona share sits 29 points from its real survey share, and the failure is uniform: every sex, age, education, and marital group sits between 24 and 34 points off, so no group is reproduced clearly better than another (Table~\ref{tab:bygroup}). The answers are also extreme. On five of the eight questions, at least one group answers within a few points of 0 or 100 (environmental means 0 to 2, work--family 100, economic support 0 to 22), where the population is split near 54 to 61. And the direction of demographic gaps is wrong: when the survey shows one group favoring a position more than another (for example, women more than men), the personas match that direction in only 34 of 110 cases.

Even the question with the smallest group-level gap does not reproduce the demographic pattern of opinion. On the one divisive question, environmental means, the personas match the survey's ordering of demographic groups in only 10 of 22 comparisons. Environmental means illustrates the mismatch: personas aged 60 and over have the lowest A-share at 32\%, whereas respondents aged 60 and over are more supportive in the human survey (53\%) than those aged 19--29 (49\%) (Table~\ref{tab:bygroup}). Thus, small differences coexist with a substantially different mapping from demographics to opinions.

\begin{table}[t]
  \centering
  \small
  \setlength{\tabcolsep}{5pt}
  \resizebox{\columnwidth}{!}{%
\begin{tabular}{lccccc}
    \toprule
    & \multicolumn{5}{c}{Share choosing Position~A (A-share, \%)} \\
    \cmidrule(lr){2-6}
    Question & Full & Demographics & Korean & No & Human \\
             & persona & only & citizen & persona & survey \\
    \midrule
    Env.\ priority   & 90 & 95 & 100 & 100 & 68 \\
    Env.\ means      & 43 & 38 & 16 & 18 & 54 \\
    Clim.\ strategy  & 97 & 89 & 19 & 1 & 70 \\
    Clim.\ tech      & 12 & 76 & 100 & 100 & 51 \\
    Work--family     & 100 & 100 & 100 & 100 & 61 \\
    Educ.--care      & 79 & 90 & 98 & 100 & 58 \\
    Econ.\ support   & 11 & 15 & 0 & 0 & 56 \\
    Housing          & 98 & 97 & 50 & 49 & 53 \\
    \midrule
    \textbf{Mean abs.\ gap (pts)} & \textbf{31} & \textbf{30} & \textbf{39} & \textbf{41} & -- \\
    \bottomrule
  \end{tabular}}
\caption{\textbf{Persona specification controls.} Persona responses as persona information is progressively removed, against the human survey.}
  \label{tab:controls}
\end{table}

\subsection{Persona conditioning shifts responses without closing the gap to the population}
\label{sec:survey-analysis}

Table~\ref{tab:controls} examines how responses change as persona information is progressively removed. With no persona conditioning, the mean absolute gap from the human survey benchmarks is 41 percentage points, and a generic ``you are a Korean citizen'' prompt leaves the gap nearly unchanged (39 points). Adding the four demographic attributes reduces the gap to 30 points, while adding the full narrative profile yields a similar gap of 31 points. Of the additions we test, the demographic attributes account for nearly all of the improvement in average alignment with the human survey benchmarks.

Changing the setup does not help either. The mismatch persists across the variations we test, including sampling at temperature 1 (mean gap 31 points; 44 of 110 directions matched). It remains under alternative response formats and when personas receive the original human questionnaire and scoring procedure, appears in an English U.S. pool benchmarked against Pew, and holds in all six models (Appendices~\ref{app:origformat} and~\ref{app:robust}): they sit 23 to 31 points from the population, and none matches the direction of demographic differences in a majority of the 110 comparisons (29 to 51). 

Together, these results show that the mismatch persists across the persona specifications, elicitation formats, populations, and models examined here.

%Changing the setup does not help either. The gap appears again in an English United States pool benchmarked against Pew and across four models including a frontier one, with full results in Appendices~\ref{app:instruments-bias}, \ref{app:origformat}, and~\ref{app:robust}. Everything varies except the model's prior, and the gap is not removable. The survey does not measure the population; it measures the model's prior, adjusted but not calibrated.
\section{Personas as deliberating agents}
\label{sec:deliberation}

A remaining question is whether interaction changes this picture: agents may respond differently once they exchange arguments with others rather than answer a policy question in isolation. We examine how their positions and arguments evolve during multi-round deliberation.

\begin{table}[t]
  \centering \small
  \begin{tabular}{lccc}
    \toprule
    Dimension (range) & Round 1 & Round 2 & Round 3 \\
    \midrule
    Justification (0--3) & 2.10 & 1.93 & 1.82 \\
    Common good (0--2)   & 1.37 & 1.34 & 1.32 \\
    Respect (0--2)       & 1.25 & 1.29 & 1.20 \\
    Reciprocity (0--1)   & 0.70 & 0.87 & 0.94 \\
    \bottomrule
  \end{tabular}
\caption{Discourse quality of GPT-6-sol debates by round: mean of two judges from other model families (Claude-Opus-5.5 and Gemini-3.1-pro; 35 rooms).}
  \label{tab:dqi}
\end{table}

\subsection{Deliberation setup}
\label{sec:delib-setup}

We follow the protocol of \citet{taubenfeld2024}, summarized in Table~\ref{tab:setup}. Stance is measured separately, by the private forced-choice survey of Section~\ref{sec:instrument}, before the debate and after each round, and is never written back into the conversation. Each room seats six agents, three on each position, who speak once per round in random order for three rounds. Two controls separate the effect of peers from other sources of change: a \emph{sealed monologue}, in which each agent sees only its own past turns, and a \emph{re-ask}, in which agents answer the same survey again without speaking; Section~\ref{sec:delib-tradeoff} adds anchored starts. Exact templates are in Appendix~\ref{app:prompts}.

\subsection{Surface features of deliberation}
\label{sec:delib-surface}

\textbf{The simulated discussions exhibit several familiar features of deliberative exchange.} Between 35 and 50\% of agents change position over the three rounds, and on six of the seven questions with balanced rooms (work--family has too few personas on Position~B) the even split resolves into a clear majority. Scored on three indicators of the Discourse Quality Index \citep{steenbergen2003} and on reciprocity \citep{graham2003,stromergalley2007} by two frontier judges from other model families (Claude-Opus-5.5 and Gemini-3.1-pro; Appendix~\ref{app:dqi})\footnote{GPT-6-sol rates its own justification at 2.75 (other judges 1.95), so it is not a judge; Appendix G adds human annotation.}, the exchanges are respectful and grow more reciprocal over the rounds (Table~\ref{tab:dqi}), and agents nearly double their distinct reasons over the debate, most giving reasons for both sides. These measures do not reveal whether agents updated: sealed monologues of the same agents score at least as high as the debates on justification, common good, and respect (Appendix~\ref{app:dqi}), and rooms in which no agent changes its mind score at least as high as live ones on every discourse measure except common good (Table~\ref{tab:frozen}).

\begin{table}[t]
\centering\small\setlength{\tabcolsep}{2.6pt}
\begin{tabular}{lcccccc}
\toprule
 & Moved & Justif. & Common & Respect & Recipr. & Both \\
Room &  & (0--3) & good & (0--2) & (0--1) & sides \\
\midrule
Live   & 47\% & 1.94 & 1.31 & 1.25 & 0.85 & 61\% \\
Frozen & 0\%  & 1.98 & 1.24 & 1.95 & 0.96 & 86\% \\
\bottomrule
\end{tabular}
\caption{Discourse quality of GPT-6-sol rooms where agents change side and rooms where none does. \emph{Moved}: agents changing side by round 3. Scores are the mean of Claude-Opus-5.5 and Gemini-3.1-pro over rounds 1--3 (Appendix~\ref{app:frozen-dqi}).}
\label{tab:frozen}
\end{table}

\begin{figure}[t]
\centering
\begin{tikzpicture}[x=0.82cm,y=0.62cm]
  \foreach \y/\name/\pop/\per/\pri/\ra/\rb/\rc/\lo/\hi in {
      6/{Env.\ priority}/4.1/5.4/6.0/6.0/5.7/6.0/5.7/6.0,
      5/{Env.\ means}/3.2/2.6/1.1/4.0/2.5/3.5/2.5/4.0,
      4/{Clim.\ strategy}/4.2/5.8/0.1/6.0/6.0/6.0/6.0/6.0,
      3/{Clim.\ tech}/3.0/0.7/6.0/4.5/1.5/0.3/0.3/4.5,
      2/{Educ.--care}/3.5/4.7/6.0/6.0/5.6/4.8/4.8/6.0,
      1/{Econ.\ support}/3.4/0.7/0.0/0.0/0.1/0.0/0.0/0.1,
      0/{Housing}/3.2/5.9/2.9/5.5/5.7/6.0/5.5/6.0}{
    \fill[gray!20] (\lo,\y-0.14) rectangle (\hi,\y+0.14);
    \draw[gray!55] (0,\y) -- (6,\y);
    \node[anchor=east,font=\small] at (-0.2,\y) {\name};
    \fill (\ra,\y) circle (1.6pt);
    \fill (\rb,\y) circle (1.6pt);
    \fill (\rc,\y) circle (1.6pt);
    \draw[line width=2.6pt] (\pri,\y-0.24) -- (\pri,\y+0.24);
    \draw[line width=2.2pt,popblue] (\pop,\y-0.17) -- (\pop,\y+0.17);
    \draw[line width=2.2pt,perorange] (\per,\y-0.17) -- (\per,\y+0.17);
  }
  \foreach \x in {0,1,2,3,4,5,6}{
    \draw[gray!55] (\x,-0.36) -- (\x,-0.26);
    \node[font=\scriptsize] at (\x,-0.65) {\x};
  }
  \node[font=\scriptsize] at (3,-1.15) {\# of agents on Position A after 3 rounds, out of 6};
  % legend, two rows below the axis
  \fill[gray!20] (-1.6,-2.02) rectangle (-1.2,-1.78);
  \fill (-1.48,-1.9) circle (1.6pt); \fill (-1.32,-1.9) circle (1.6pt);
  \node[font=\scriptsize,anchor=west] at (-1.1,-1.9) {where rooms end (three starts)};
  \draw[line width=2.6pt] (3.0,-2.1) -- (3.0,-1.7);
  \node[font=\scriptsize,anchor=west] at (3.15,-1.9) {prior (no persona)};
  \draw[line width=2.2pt,popblue] (-1.4,-2.75) -- (-1.4,-2.4);
  \node[font=\scriptsize,anchor=west] at (-1.1,-2.58) {population};
  \draw[line width=2.2pt,perorange] (3.0,-2.75) -- (3.0,-2.4);
  \node[font=\scriptsize,anchor=west] at (3.15,-2.58) {personas' survey};
\end{tikzpicture}
\caption{Where rooms end against the three references. The shaded span covers the mean endpoints of
rooms starting all-A, balanced, and all-B (dots). The population lies outside the span on five of the seven questions; other models in Appendix~\ref{app:composition}.}
\label{fig:composition}
\end{figure}

\paragraph{Where rooms end depends little on where they start, and not on the population.} On six of the seven questions, rooms starting all-A, balanced, or all-B end close together, sometimes at a majority no agent held at the start (Figure~\ref{fig:composition}). That endpoint is not the population's: the population lies outside it on five of the seven questions, and it sits on the side most personas chose in the survey.

\paragraph{Much of the observed movement does not require peer interaction.} Rerun as sealed monologues, the rooms end in nearly the same place (Figure~\ref{fig:mono}); in five of six models the difference lies within half the smallest change a single room can show (Appendix~\ref{app:graded}). Peers leave a trace room by room (0.32 agents per room beyond rerun noise; Appendix~\ref{app:graded}), mostly on environmental means, the one divisive question, and re-asking the survey without speaking changes 16\% of answers (5--30\% across models) but leaves rooms near the even start. Much of the observed stance movement can thus arise without hearing other agents.

\begin{figure}[t]
\centering
\begin{tikzpicture}[x=0.8cm,y=0.8cm]
\foreach \v in {0,...,6}{
  \draw[gray!22] (\v,0) -- (\v,6); \draw[gray!22] (0,\v) -- (6,\v);
  \node[font=\tiny,below] at (\v,0) {\v}; \node[font=\tiny,left] at (0,\v) {\v};}
\draw[gray!70] (0,0) rectangle (6,6);
\draw[black!55,dashed] (0,0) -- (6,6);
\node[font=\scriptsize] at (3,-0.62) {Monologue (agents on Position~A of 6)};
\node[font=\scriptsize,rotate=90] at (-0.62,3) {Debate (agents on Position~A of 6)};
\foreach \x/\y in {5.85/6.00,4.30/4.15,5.95/6.00,5.30/6.00,0.00/0.00,5.25/5.30,5.80/5.80,3.65/2.60,4.60/5.05,0.90/1.10,4.15/3.90,4.70/4.60,5.50/5.30,0.55/0.50,4.40/4.50,4.50/5.60,1.65/0.55,3.05/4.05,5.15/6.00,5.95/6.00,1.65/1.65,5.05/5.55,5.00/5.20,0.50/0.00,2.80/3.25,5.55/5.05,0.90/1.60,3.85/4.75,0.90/1.05,2.40/3.00}
  \draw[draw=gray!80,fill=white,line width=0.5pt] (\x,\y) circle (2.2pt);
\foreach \x/\y in {1.85/1.35,2.10/3.15,1.80/2.80,3.30/3.00,1.80/2.45,1.95/2.50}
  \draw[draw=perorange,fill=perorange!45,line width=0.5pt] (\x,\y) circle (2.2pt);
\foreach \x/\y in {5.70/5.70,6.00/5.95,1.45/1.50,5.80/5.55,0.05/0.05,5.55/5.70}
  \draw[draw=popblue,fill=popblue,line width=0.5pt] (\x,\y) circle (2.2pt);
\begin{scope}[shift={(2.95,1.05)}]
\fill[white,opacity=0.9] (-0.2,-0.95) rectangle (2.55,0.25);
\draw[draw=popblue,fill=popblue] (0,0) circle (2.2pt); \node[font=\tiny,anchor=west] at (0.12,0) {GPT-6-sol};
\draw[draw=gray!80,fill=white] (0,-0.35) circle (2.2pt); \node[font=\tiny,anchor=west] at (0.12,-0.35) {Other five models};
\draw[draw=perorange,fill=perorange!45] (0,-0.7) circle (2.2pt); \node[font=\tiny,anchor=west] at (0.12,-0.7) {Environmental means (divisive)};
\end{scope}
\end{tikzpicture}
\caption{Final room composition in debate against the sealed monologue of the same agents, one point per model and question (balanced rooms). Points on the dashed line end in the same place either way.}
\label{fig:mono}
\end{figure}

\begin{figure}[t]
\centering
\begin{tikzpicture}[x=1cm,y=0.04cm]
\foreach \y in {0,20,40,60}{\draw[gray!22] (0,\y) -- (6.6,\y); \node[font=\tiny,left] at (0,\y) {\y};}
\draw[gray!70] (0,0) -- (6.6,0);
\node[font=\scriptsize,rotate=90] at (-0.62,30) {Agents changing side (\%)};
\node[font=\scriptsize,below] at (1.1,0) {Nothing restated};
\node[font=\scriptsize,below] at (3.3,0) {Position restated};
\node[font=\scriptsize,below] at (5.5,0) {Opening argued};
% bars: GPT-6-sol (center / height / shade)
\foreach \x/\h/\s in {0.77/48.3/35,1.43/50.3/12,2.97/0.5/35,3.63/1.0/12,5.17/0.5/35,5.83/0.5/12}{
  \fill[gray!\s] (\x-0.28,0) rectangle (\x+0.28,\h); \draw[gray!60] (\x-0.28,0) rectangle (\x+0.28,\h);}
% dots: six models, same order in every bar
\foreach \x/\a/\b/\c/\d/\e/\f in {0.77/48.3/48.5/39.6/44.4/45.7/43.5, 1.43/50.3/49.3/49.0/50.0/49.3/48.3,
  2.97/0.0/14.7/1.4/8.7/10.6/5.1, 3.63/1.0/21.9/1.4/14.6/26.7/8.3,
  5.17/0.0/11.8/1.4/3.5/13.5/8.3, 5.83/0.0/19.8/3.1/9.4/21.5/9.4}{
  \foreach \o/\v/\m in {-0.2/\a/msol,-0.12/\b/mluna,-0.04/\c/mmini,0.04/\d/mds,0.12/\e/mhaiku,0.2/\f/mqwen}
    \fill[\m,draw=white,line width=0.3pt] (\x+\o,\v) circle (2.1pt);}
% legend
\fill[gray!35] (3.7,58.6) rectangle (3.94,63.4); \draw[gray!60] (3.7,58.6) rectangle (3.94,63.4); \node[font=\scriptsize,anchor=west] at (4.0,61) {Survey-chosen};
\fill[gray!12] (3.7,50.6) rectangle (3.94,55.4); \draw[gray!60] (3.7,50.6) rectangle (3.94,55.4); \node[font=\scriptsize,anchor=west] at (4.0,53) {Population-matched};
\foreach \x/\y/\m/\n in {0.15/-20/msol/GPT-6-sol,2.35/-20/mluna/GPT-6-luna,4.55/-20/mmini/GPT-4.1-mini,
  0.15/-27/mds/DeepSeek-V4.1,2.35/-27/mhaiku/Claude-Haiku-5.5,4.55/-27/mqwen/Qwen-3.8-flash}{
  \fill[\m,draw=white,line width=0.3pt] (\x,\y) circle (2.1pt); \node[font=\tiny,anchor=west] at (\x+0.08,\y) {\n};}
\end{tikzpicture}
\caption{Agents changing side over three rounds, by how the starting position is assigned (bar shade) and carried into the debate. Bars: GPT-6-sol; dots: each of the six models (Table~\ref{tab:grid6}). Restating the position sharply reduces movement under either assignment.}
\label{fig:anchor}
\end{figure}

\subsection{When a starting position is restated, reported stances change far less}
\label{sec:delib-tradeoff}

Figure~\ref{fig:anchor} gives each agent a starting position, either its own survey answer (\emph{survey-chosen}) or one drawn from the population benchmarks with the room kept balanced (\emph{population-matched}), and leaves it unstated, restates it in one line, or has the agent argue it in an opening turn. Left unstated, about half the agents change side under either assignment. Restated or argued, movement falls in every model (Table~\ref{tab:grid6}): a one-line restatement anchors as strongly as an argued opening. The anchor runs through the agent's own turns: removing them from the stance-survey context restores movement, whereas removing a same-side peer's turns does not (Appendix~\ref{app:inject}).

In sum, these simulations produce reciprocal, both-sided exchanges with substantial stance movement, but much of the movement can arise without peers, and restating an assigned position sharply reduces later changes. Argument generation, population representation, and interaction-driven opinion change are thus distinct properties of LLM deliberation.
\section{Testing against a real deliberative poll}
\label{sec:poll}

\begin{figure*}[t]
\centering
\begin{tikzpicture}[x=0.42cm,y=0.028cm]
\begin{scope}[xshift=0.00cm]
\draw[gray!25] (0,20) -- (4,20);
\draw[gray!25] (0,40) -- (4,40);
\draw[gray!25] (0,60) -- (4,60);
\draw[gray!25] (0,80) -- (4,80);
\draw[gray!60] (0,20) -- (0,95);
\draw[gray!60] (0,20) -- (4,20);
\node[font=\tiny,below] at (0,20) {W1};
\node[font=\tiny,below] at (1,20) {Bk};
\node[font=\tiny,below] at (2,20) {R1};
\node[font=\tiny,below] at (3,20) {R2};
\node[font=\tiny,below] at (4,20) {R3};
\node[font=\tiny,left] at (0,20) {20};
\node[font=\tiny,left] at (0,40) {40};
\node[font=\tiny,left] at (0,60) {60};
\node[font=\tiny,left] at (0,80) {80};
\node[font=\scriptsize,rotate=90] at (-1.55,57.5) {Resume (\%)};
\node[font=\scriptsize] at (2,101) {GPT-6-sol};
\draw[trdeb,line width=0.9pt] (0,47.0) -- (1,41.0) -- (2,44.2) -- (3,44.2) -- (4,44.2);
\draw[trmono,line width=0.9pt,densely dashed] (0,46.0) -- (1,41.8) -- (2,43.6) -- (3,45.4) -- (4,45.8);
\draw[trrest,line width=1pt,densely dotted] (0,48.2) -- (1,42.0) -- (2,51.4) -- (3,51.4) -- (4,51.4);
\draw[black,line width=1.2pt] (0,57.0) -- (1,59.3) -- (4,59.2);
\foreach \p in {(0,57.0),(1,59.3),(4,59.2)} \fill[black] \p circle (1.6pt);
\end{scope}
\begin{scope}[xshift=2.62cm]
\draw[gray!25] (0,20) -- (4,20);
\draw[gray!25] (0,40) -- (4,40);
\draw[gray!25] (0,60) -- (4,60);
\draw[gray!25] (0,80) -- (4,80);
\draw[gray!60] (0,20) -- (0,95);
\draw[gray!60] (0,20) -- (4,20);
\node[font=\tiny,below] at (0,20) {W1};
\node[font=\tiny,below] at (1,20) {Bk};
\node[font=\tiny,below] at (2,20) {R1};
\node[font=\tiny,below] at (3,20) {R2};
\node[font=\tiny,below] at (4,20) {R3};
\node[font=\scriptsize] at (2,101) {GPT-6-luna};
\draw[trdeb,line width=0.9pt] (0,69.5) -- (1,63.3) -- (2,45.0) -- (3,45.0) -- (4,45.0);
\draw[trmono,line width=0.9pt,densely dashed] (0,73.9) -- (1,68.1) -- (2,49.2) -- (3,48.2) -- (4,47.6);
\draw[trrest,line width=1pt,densely dotted] (0,73.1) -- (1,63.1) -- (2,53.2) -- (3,53.0) -- (4,52.8);
\draw[black,line width=1.2pt] (0,57.0) -- (1,59.3) -- (4,59.2);
\foreach \p in {(0,57.0),(1,59.3),(4,59.2)} \fill[black] \p circle (1.6pt);
\end{scope}
\begin{scope}[xshift=5.24cm]
\draw[gray!25] (0,20) -- (4,20);
\draw[gray!25] (0,40) -- (4,40);
\draw[gray!25] (0,60) -- (4,60);
\draw[gray!25] (0,80) -- (4,80);
\draw[gray!60] (0,20) -- (0,95);
\draw[gray!60] (0,20) -- (4,20);
\node[font=\tiny,below] at (0,20) {W1};
\node[font=\tiny,below] at (1,20) {Bk};
\node[font=\tiny,below] at (2,20) {R1};
\node[font=\tiny,below] at (3,20) {R2};
\node[font=\tiny,below] at (4,20) {R3};
\node[font=\scriptsize] at (2,101) {GPT-4.1-mini};
\draw[trdeb,line width=0.9pt] (0,43.2) -- (1,38.6) -- (2,49.0) -- (3,48.6) -- (4,49.1);
\draw[trmono,line width=0.9pt,densely dashed] (0,44.2) -- (1,38.4) -- (2,37.6) -- (3,45.0) -- (4,47.8);
\draw[trrest,line width=1pt,densely dotted] (0,44.0) -- (1,37.8) -- (2,54.0) -- (3,53.6) -- (4,53.8);
\draw[black,line width=1.2pt] (0,57.0) -- (1,59.3) -- (4,59.2);
\foreach \p in {(0,57.0),(1,59.3),(4,59.2)} \fill[black] \p circle (1.6pt);
\end{scope}
\begin{scope}[xshift=7.86cm]
\draw[gray!25] (0,20) -- (4,20);
\draw[gray!25] (0,40) -- (4,40);
\draw[gray!25] (0,60) -- (4,60);
\draw[gray!25] (0,80) -- (4,80);
\draw[gray!60] (0,20) -- (0,95);
\draw[gray!60] (0,20) -- (4,20);
\node[font=\tiny,below] at (0,20) {W1};
\node[font=\tiny,below] at (1,20) {Bk};
\node[font=\tiny,below] at (2,20) {R1};
\node[font=\tiny,below] at (3,20) {R2};
\node[font=\tiny,below] at (4,20) {R3};
\node[font=\scriptsize] at (2,101) {DeepSeek-V4.1};
\draw[trdeb,line width=0.9pt] (0,28.7) -- (1,34.0) -- (2,31.4) -- (3,31.1) -- (4,30.6);
\draw[trmono,line width=0.9pt,densely dashed] (0,30.5) -- (1,31.1) -- (2,28.7) -- (3,27.0) -- (4,27.9);
\draw[trrest,line width=1pt,densely dotted] (0,29.6) -- (1,31.5) -- (2,44.4) -- (3,44.0) -- (4,43.2);
\draw[black,line width=1.2pt] (0,57.0) -- (1,59.3) -- (4,59.2);
\foreach \p in {(0,57.0),(1,59.3),(4,59.2)} \fill[black] \p circle (1.6pt);
\end{scope}
\begin{scope}[xshift=10.48cm]
\draw[gray!25] (0,20) -- (4,20);
\draw[gray!25] (0,40) -- (4,40);
\draw[gray!25] (0,60) -- (4,60);
\draw[gray!25] (0,80) -- (4,80);
\draw[gray!60] (0,20) -- (0,95);
\draw[gray!60] (0,20) -- (4,20);
\node[font=\tiny,below] at (0,20) {W1};
\node[font=\tiny,below] at (1,20) {Bk};
\node[font=\tiny,below] at (2,20) {R1};
\node[font=\tiny,below] at (3,20) {R2};
\node[font=\tiny,below] at (4,20) {R3};
\node[font=\scriptsize] at (2,101) {Claude-Haiku-5.5};
\draw[trdeb,line width=0.9pt] (0,51.4) -- (1,34.7) -- (2,41.4) -- (3,43.0) -- (4,36.9);
\draw[trmono,line width=0.9pt,densely dashed] (0,51.2) -- (1,35.5) -- (2,45.0) -- (3,46.6) -- (4,45.6);
\draw[trrest,line width=1pt,densely dotted] (0,51.6) -- (1,35.1) -- (2,48.0) -- (3,49.4) -- (4,51.8);
\draw[black,line width=1.2pt] (0,57.0) -- (1,59.3) -- (4,59.2);
\foreach \p in {(0,57.0),(1,59.3),(4,59.2)} \fill[black] \p circle (1.6pt);
\end{scope}
\begin{scope}[xshift=13.10cm]
\draw[gray!25] (0,20) -- (4,20);
\draw[gray!25] (0,40) -- (4,40);
\draw[gray!25] (0,60) -- (4,60);
\draw[gray!25] (0,80) -- (4,80);
\draw[gray!60] (0,20) -- (0,95);
\draw[gray!60] (0,20) -- (4,20);
\node[font=\tiny,below] at (0,20) {W1};
\node[font=\tiny,below] at (1,20) {Bk};
\node[font=\tiny,below] at (2,20) {R1};
\node[font=\tiny,below] at (3,20) {R2};
\node[font=\tiny,below] at (4,20) {R3};
\node[font=\scriptsize] at (2,101) {Qwen-3.8-flash};
\draw[trdeb,line width=0.9pt] (0,89.0) -- (1,61.4) -- (2,46.8) -- (3,44.8) -- (4,43.0);
\draw[trmono,line width=0.9pt,densely dashed] (0,89.4) -- (1,60.2) -- (2,50.8) -- (3,54.0) -- (4,48.6);
\draw[trrest,line width=1pt,densely dotted] (0,88.0) -- (1,61.8) -- (2,53.4) -- (3,52.8) -- (4,52.8);
\draw[black,line width=1.2pt] (0,57.0) -- (1,59.3) -- (4,59.2);
\foreach \p in {(0,57.0),(1,59.3),(4,59.2)} \fill[black] \p circle (1.6pt);
\end{scope}
% legend
\begin{scope}[yshift=-0.78cm]
\draw[trdeb,line width=0.9pt] (8.6,20) -- (9.6,20); \node[font=\scriptsize,anchor=west] at (9.7,20) {Debate};
\draw[trmono,line width=0.9pt,densely dashed] (12.4,20) -- (13.4,20); \node[font=\scriptsize,anchor=west] at (13.5,20) {Monologue};
\draw[trrest,line width=1pt,densely dotted] (17.0,20) -- (18.0,20); \node[font=\scriptsize,anchor=west] at (18.1,20) {Restated};
\draw[black,line width=1.2pt] (21.0,20) -- (22.0,20); \fill[black] (21.5,20) circle (1.6pt); \node[font=\scriptsize,anchor=west] at (22.1,20) {Real panel};
\end{scope}
\end{tikzpicture}
\caption{Share choosing \emph{resume} among agents choosing a side, by stage (W1: wave 1; Bk: after the summaries; R1--R3: after each round), for debate, sealed monologue, and restated debate, against the real panel at waves 1, 3, and 4.}
\label{fig:traj}
\end{figure*}

Even when agents are built from real survey respondents' records, they fail to reproduce both the panel's starting opinions and its subsequent changes.
Simulated deliberation also does not consistently improve agreement with the real panel over sealed monologue.
 
\subsection{Poll setup}
\label{sec:poll-setup}\label{sec:shinkori}

The 2017 Shin-Kori deliberative poll asked whether to resume or halt construction of two nuclear reactors in South Korea \citep{shingori2017}.
We use three survey waves: before information (wave~1), after presentations and study materials (wave~3), and after a three-day deliberative session (wave~4).

For each model, we build 498 agents from first-wave survey records.
We sample these records to match the panel's joint distribution of initial opinion, sex, and age.
Each agent is created using the respondent's age, sex, province, occupation, and party affiliation.

We follow the survey and deliberation protocol of Sections~\ref{sec:respondents} and~\ref{sec:deliberation}, with debates in 83 groups of six.
Before discussion, agents read short summaries of both sides' arguments and answer the survey again.
We also run the sealed-monologue and restated-position conditions with the same agents.

Our measure is the \emph{resume share}: the percentage choosing \emph{resume} among those choosing either \emph{resume} or \emph{halt}.
We compare the agents' initial, post-information, and final shares with the panel's corresponding survey waves (Appendix~\ref{app:poll}).

\begin{table}[t]
\centering\small
\setlength{\tabcolsep}{3.2pt}
\resizebox{\columnwidth}{!}{%
\begin{tabular}{l cc ccc}
\toprule
& \multicolumn{2}{c}{Survey stage}
& \multicolumn{3}{c}{After three rounds} \\
\cmidrule(lr){2-3}\cmidrule(lr){4-6}
& Wave 1 & Booklet & Debate & Monologue & Restated \\
\midrule
\textbf{Real panel}
& 57.0 & 59.3 & 59.2 & 59.2 & 59.2 \\
\midrule
GPT-6-sol
& 47.0 & 41.0 & 44.2 & 45.8 & 51.4 \\
GPT-6-luna
& 69.5 & \textbf{63.3} & 45.0 & 47.6 & \textbf{52.8} \\
GPT-4.1-mini
& 43.2 & 38.6 & 49.1 & 47.8 & \textbf{53.8} \\
DeepSeek-V4.1
& 28.7 & 34.0 & 30.6 & 27.9 & 43.2 \\
Claude-Haiku-5.5
& \textbf{51.4} & 34.7 & 36.9 & 45.6 & 51.8 \\
Qwen-3.8-flash
& 89.0 & \textbf{61.4} & 43.0 & 48.6 & 52.8 \\
\bottomrule
\end{tabular}%
}
\caption{Resume share (\%) by stage and discussion condition.
Bold values indicate that the 95\% interval for the difference from the real panel includes zero (Appendix~\ref{app:poll}).}
\label{tab:poll}
\end{table}

\subsection{Agents do not reproduce the panel's starting opinions}
\label{sec:poll-survey}

Before receiving information, agents' resume shares range from 29\% to 89\%, compared with 57\% in the real panel (Table~\ref{tab:poll}).
Five of the six models differ significantly from the panel.
When two demographic or political groups differ by more than five percentage points in the real survey, models usually identify which group is more supportive of resuming construction.
However, their average error in group-level resume shares remains 10--32 percentage points, depending on the model (Appendix~\ref{app:poll}).

After the information stage, the panel's resume share rises from 57.0\% to 59.3\%.
In contrast, reading our summaries lowers the resume share in five of six models, by up to 28 percentage points.
GPT-6-luna's share falls from 69.5\% to 63.3\%, and Qwen-3.8-flash's falls from 89.0\% to 61.4\%.
Both become closer to the panel's 59.3\%, but their support for resuming decreases while the panel's increases.

\subsection{Simulated deliberation does not reproduce the panel's opinion changes}
\label{sec:poll-delib}

During the real deliberative session, the panel's resume share barely changed, from 59.3\% to 59.2\%.
In contrast, simulated deliberation changed resume shares by up to 18 percentage points, and all six models ended with a majority favoring halting construction (Table~\ref{tab:poll}; Figure~\ref{fig:traj}).
Sealed monologues produced similar results, with final resume shares differing from simulated deliberation by at most 9 percentage points.
Neither condition reproduced the panel's final resume share in any model (Appendix~\ref{app:poll}). Telling agents their respondent's real wave-1 answer brings final shares closer to the panel. This is mostly because agents take the side they are told (Appendix~\ref{app:poll}).

\paragraph{Agents discount what their peers say.}
We test whether the source of an argument affects agents' opinions.
Each agent receives the same argument against its post-information position, but with a different source label: another participant, an expert, the agent's own previous turn, or no source.
In every model, fewer agents switch sides when the argument is attributed to another participant than when it has no source.
The peer label also produces fewer switches than the expert label in five of six models.
In contrast, 81--93\% of agents switch sides when the argument is attributed to their own previous turn (Figure~\ref{fig:attrib}; Appendix~\ref{app:poll}).

\begin{figure}[t]
\centering
\begin{tikzpicture}[x=1.55cm,y=0.024cm]
\foreach \y in {0,25,50,75,100}{\draw[gray!22] (0.45,\y) -- (4.55,\y); \node[font=\tiny,left] at (0.45,\y) {\y};}
\draw[gray!70] (0.45,0) -- (4.55,0);
\node[font=\scriptsize,rotate=90] at (0.08,50) {Agents moved (\%)};
\fill[gray!30] (0.82,0) rectangle (1.18,92.6); \draw[gray!70] (0.82,0) rectangle (1.18,92.6);
\foreach \o/\v/\c in {-0.13/92.6/msol,-0.078/91.8/mluna,-0.026/90.6/mmini,0.026/82.7/mds,0.078/80.7/mhaiku,0.13/84.9/mqwen} \fill[\c,draw=white,line width=0.3pt] (1+\o,\v) circle (1.9pt);
\fill[gray!30] (1.82,0) rectangle (2.18,74.5); \draw[gray!70] (1.82,0) rectangle (2.18,74.5);
\foreach \o/\v/\c in {-0.13/74.5/msol,-0.078/91.0/mluna,-0.026/43.0/mmini,0.026/37.3/mds,0.078/36.1/mhaiku,0.13/42.4/mqwen} \fill[\c,draw=white,line width=0.3pt] (2+\o,\v) circle (1.9pt);
\fill[gray!30] (2.82,0) rectangle (3.18,33.5); \draw[gray!70] (2.82,0) rectangle (3.18,33.5);
\foreach \o/\v/\c in {-0.13/33.5/msol,-0.078/76.3/mluna,-0.026/41.6/mmini,0.026/21.5/mds,0.078/14.7/mhaiku,0.13/16.3/mqwen} \fill[\c,draw=white,line width=0.3pt] (3+\o,\v) circle (1.9pt);
\fill[gray!30] (3.82,0) rectangle (4.18,18.1); \draw[gray!70] (3.82,0) rectangle (4.18,18.1);
\foreach \o/\v/\c in {-0.13/18.1/msol,-0.078/57.0/mluna,-0.026/19.5/mmini,0.026/9.8/mds,0.078/18.7/mhaiku,0.13/11.2/mqwen} \fill[\c,draw=white,line width=0.3pt] (4+\o,\v) circle (1.9pt);
\node[font=\scriptsize,below,align=center] at (1,0) {Own last\\turn};
\node[font=\scriptsize,below,align=center] at (2,0) {No\\source};
\node[font=\scriptsize,below] at (3,0) {Expert};
\node[font=\scriptsize,below,align=center] at (4,0) {Fellow\\participant};
\fill[msol,draw=white,line width=0.3pt] (0.55,-45) circle (1.9pt); \node[font=\tiny,anchor=west] at (0.63,-45) {GPT-6-sol};
\fill[mluna,draw=white,line width=0.3pt] (1.97,-45) circle (1.9pt); \node[font=\tiny,anchor=west] at (2.05,-45) {GPT-6-luna};
\fill[mmini,draw=white,line width=0.3pt] (3.39,-45) circle (1.9pt); \node[font=\tiny,anchor=west] at (3.47,-45) {GPT-4.1-mini};
\fill[mds,draw=white,line width=0.3pt] (0.55,-58) circle (1.9pt); \node[font=\tiny,anchor=west] at (0.63,-58) {DeepSeek-V4.1};
\fill[mhaiku,draw=white,line width=0.3pt] (1.97,-58) circle (1.9pt); \node[font=\tiny,anchor=west] at (2.05,-58) {Claude-Haiku-5.5};
\fill[mqwen,draw=white,line width=0.3pt] (3.39,-58) circle (1.9pt); \node[font=\tiny,anchor=west] at (3.47,-58) {Qwen-3.8-flash};
\end{tikzpicture}
\caption{The same counter-argument, by the source it is attributed to (Shin-Kori agents at the booklet stage, 498 per model). Bars: GPT-6-sol; dots: each of the six models. Agents follow a statement most when it is labeled as their own and, in five of six models, least when it comes from a fellow participant (Appendix~\ref{app:poll}).}
\label{fig:attrib}
\end{figure}

\paragraph{Knowing the real outcome does not explain the mismatch.}
All six models correctly identify that the real panel recommended resuming construction when asked directly without a persona.
To test whether this knowledge affects the simulation, we repeat the experiment using a fictional plant name.
No model comes more than 2 percentage points closer to the real panel, and three models move further away (Appendix~\ref{app:poll}).

Overall, agents built from real survey records neither reproduce the panel's initial opinions nor its changes across survey waves.
Simulated deliberation produces results similar to sealed monologue, and agents are less likely to follow arguments attributed to peers than the same arguments without a source.
Thus, reproducing the stages of a deliberative poll does not necessarily reproduce how the panel's opinions change.

\section{Discussion and Future Directions}
\label{sec:discussion}

\subsection{Population simulation}
Our results caution against using LLM deliberation as a population simulation. The agents do not reliably reproduce baseline opinion patterns, much of their observed stance movement can arise without peer interaction, and agents built from real participants do not change as a real deliberating public did. A reasoned and reciprocal transcript therefore does not by itself show that the simulation captures how a population would deliberate or where its opinions would move. A different use remains more open: using LLM deliberation to surface arguments and social perspectives around a contested question. Our agents generate varied, both-sided arguments and expand their argument repertoires over the discussion. These findings suggest a potentially useful role for LLM deliberation in surfacing arguments and perspectives for human inspection and reflection.

\subsection{Argument surfacing}
If the argument-surfacing use can be validated, a further challenge is how to make the resulting range of arguments and perspectives useful to human readers. 
%If LLM deliberation can reliably surface a diverse range of arguments and perspectives, its value need not lie only in predicting a collective outcome. 
Such a use could let people explore the positions and reasons that arise around a question, as well as how they develop through reason exchange, an important part of deliberation beyond its outcome \citep{bachtiger2018}. Such exploration is difficult because each debate yields a long interaction history, while simulation runs across varying conditions produce many possible trajectories.

\subsection{Visualizing simulated deliberation}
% Visualization offers one way to address this challenge. The challenges of interpreting simulated deliberations motivate three design requirements: \emph{overview}, \emph{traceability}, and \emph{comparability}. First, because deliberations produce long interaction histories, users need an \emph{overview} that shows the distribution of agents’ positions and reasons without requiring them to read full transcripts. Second, since understanding the process is essential in political deliberation, the development of individual views and reasons should be \emph{traceable}. Finally, as multi-agent simulation can generate alternative deliberative trajectories across discussion conditions or information interventions, these paths should be \emph{comparable}. Together, these views suggest a complementary use of LLM deliberation: not simply to produce one predicted answer, but to support human reflection on diverse reasons, judgment formation, and the conditions that may shape collective opinion.
Visualization offers one way to address this challenge, through three design requirements: an \emph{overview} of agents' positions and reasons that does not require reading full transcripts, \emph{traceability} of how individual views and reasons develop, and \emph{comparability} of deliberative trajectories across discussion conditions or information interventions.

\subsection*{Concluding remarks}
More broadly, the promise of LLM deliberation depends on distinguishing what a simulation is intended to represent and validating it accordingly. Simulating population opinion dynamics requires evidence that agents represent the population and respond to one another, while using deliberation to surface social perspectives requires evidence that the generated arguments reflect those perspectives. Developing and validating both uses could make LLM deliberation a more reliable tool for understanding public reasoning.

\section*{Limitations}

Our available benchmarks measure opinions, not arguments or perspectives, so we cannot evaluate argument-surfacing directly. This highlights the need for distinct validation benchmarks for population representation, interaction-driven updating, and the diversity of human perspectives.

Other limitations concern scope and measurement. Deliberation dynamics are tested with six models and three policy domains in Korea. The U.S. evidence is limited to the survey stage, and room-level estimates are based on six-agent debates. Future work should test whether the patterns observed here persist across models, populations, policy domains, and deliberation designs.

Removing an agent's own prior turns from the stance-survey context restores substantial movement, and in the deliberative poll a statement moves agents most when it is labeled as their own, suggesting that its own discourse history plays an important role in subsequent stance reporting. However, we do not distinguish among possible mechanisms, such as self-consistency, self-persuasion from repeatedly generating reasons, context accumulation, reading a restated position as an instruction, or other prompt-induced dynamics. Distinguishing among these explanations will require more targeted interventions on agents' own discourse histories.

In Section 6, the discussion stage is thinner than the real three-day session, which also had expert presentations, question-and-answer sessions, and moderators, so it tests whether agents reproduce the effect of the session as a whole. The panel was surveyed at fixed points and later-wave microdata were not released, so we compare each wave's distribution, not individual trajectories. Each agent's persona holds only the five attributes of the first-wave record.

% Acknowledgments are omitted in the anonymized review version.
%\section*{Acknowledgments}

\section*{Ethics Statement}

The study protocol was approved by our institution's IRB. The Shin-Kori first-wave survey data were released by the deliberation committee without names or contact information. We use only the released variables (age, sex, province, occupation, party, and the wave-1 answer), and the outputs we release contain only these profile lines and generated text, which contain no names, contact information, or record identifiers. The KEI microdata are publicly distributed for research use. Simulated participants are not people, and our results argue against using them to stand in for a public in policy decisions.


\bibliography{custom}

\renewcommand{\topfraction}{0.95}\renewcommand{\bottomfraction}{0.8}\renewcommand{\textfraction}{0.05}
\renewcommand{\floatpagefraction}{0.85}\renewcommand{\dbltopfraction}{0.95}\renewcommand{\dblfloatpagefraction}{0.85}
\setcounter{topnumber}{4}\setcounter{bottomnumber}{3}\setcounter{totalnumber}{6}\setcounter{dbltopnumber}{3}
\appendix
\section{Question wordings and by-group results}
\label{app:instruments}
\label{app:bygroup}
Wordings are translated from the Korean source items. KEI is the Korea Environment Institute's 2025 National Environmental Awareness Survey, and PCASPP the low-birthrate perception survey of the Presidential Committee on Ageing Society and Population Policy; the Pew sources of the U.S. items are listed in the released code. The low-birthrate survey covers adults aged 20--49; its age groups use the 345 personas in that range, and restricting every low-birthrate comparison to them changes each question's A-share by at most 4 points. Table~\ref{tab:bygroup} gives GPT-6-sol persona shares against the survey by group. The divisive/saturated classification of Section~\ref{sec:instrument} is unchanged under a 70\% cutoff.

\begin{enumerate}\itemsep2pt
\item \textbf{Environmental priority} (\emph{saturated}; persona share 90\%, population 68\%). \emph{Topic:} Environmental policy priority. \emph{Position~1 (A):} Environmental protection first (even if economic growth slows somewhat). \emph{Position~2 (B):} Economic growth first (even if the environment is somewhat harmed).
\item \textbf{Environmental means} (\emph{divisive}; persona share 43\%, population 54\%). \emph{Topic:} Means of solving environmental problems. \emph{Position~1 (A):} Stronger penalties and law enforcement. \emph{Position~2 (B):} Voluntary efforts by citizens and firms first.
\item \textbf{Climate strategy} (\emph{saturated}; persona share 97\%, population 70\%). \emph{Topic:} Climate-crisis response strategy (priority use of limited resources). \emph{Position~1 (A):} Climate adaptation first (infrastructure for floods and heat waves). \emph{Position~2 (B):} Energy transition first (expanding renewables).
\item \textbf{Climate technology} (\emph{saturated}; persona share 12\%, population 51\%). \emph{Topic:} Priority for fostering climate technology. \emph{Position~1 (A):} Renewable-energy technology (solar and wind). \emph{Position~2 (B):} Circular-economy technology (recycling and waste management).
\item \textbf{Work--family} (\emph{saturated}; persona share 100\%, population 61\%). \emph{Topic:} Work--family balance policy priority. \emph{Position~1 (A):} Promoting flexible work during child-rearing. \emph{Position~2 (B):} Further raising the parental-leave benefit cap.
\item \textbf{Education--care} (\emph{saturated}; persona share 79\%, population 58\%). \emph{Topic:} Education and care policy priority. \emph{Position~1 (A):} Expanding government support for childcare services (coverage and hours). \emph{Position~2 (B):} Improving curriculum and content to reduce private-education costs.
\item \textbf{Economic support} (\emph{saturated}; persona share 11\%, population 56\%). \emph{Topic:} Form of economic support for marriage and childbirth. \emph{Position~1 (A):} Expanding tax benefits for married and child-rearing households. \emph{Position~2 (B):} Expanding cash support (parental and child allowances).
\item \textbf{Housing} (\emph{saturated}; persona share 98\%, population 53\%). \emph{Topic:} Direction of housing support for low birthrate. \emph{Position~1 (A):} Loosening income thresholds for home-purchase and jeonse loans. \emph{Position~2 (B):} Expanding housing-subscription special provisions for newlywed and child-rearing households.
\end{enumerate}

\newcolumntype{Y}{>{\centering\arraybackslash}X}

\begin{table*}[!tbp]
%Each issue cell reports the A-share as \textbf{Full persona / Human survey} (\%). The final column reports the mean absolute gap across issues for each group; the bottom row reports the mean absolute gap across groups for each issue.}
\centering
\small
\setlength{\tabcolsep}{4pt}
% ---------- Panel A ----------
\begin{tabularx}{\textwidth}{@{}p{1.55cm}p{1.65cm}YYYYY@{}}
\multicolumn{7}{@{}l}{}\\
%\textbf{(a) Environmental policy}} \\[2pt]
\toprule
\multicolumn{2}{l}{}
\textbf{(a) Environmental policy}
& Env.\ priority
& Env.\ means
& Clim.\ strategy
& Clim.\ tech
& \textbf{Gap} \\
\midrule
\multirow{2}{=}{Sex}
& Male
& 83 / 57 & 49 / 54 & 98 / 67 & 13 / 53 & \textbf{25} \\
& Female
& 97 / 80 & 37 / 54 & 97 / 74 & 11 / 48 & \textbf{24} \\
\addlinespace[3pt]
\multirow{4}{=}{Age}
& 19--29
& 83 / 51 & 39 / 49 & 97 / 74 & 19 / 47 & \textbf{23} \\
& 30--44
& 87 / 58 & 50 / 57 & 96 / 71 & 13 / 53 & \textbf{25} \\
& 45--59
& 93 / 75 & 52 / 55 & 98 / 69 & 8 / 50 & \textbf{23} \\
& 60+
& 96 / 81 & 32 / 53 & 99 / 69 & 9 / 51 & \textbf{27} \\
\addlinespace[3pt]
\multirow{4}{=}{Education}
& $\leq$ Middle
& 86 / 87 & 42 / 57 & 100 / 50 & 3 / 64 & \textbf{32} \\
& High school
& 84 / 73 & 40 / 53 & 100 / 70 & 6 / 48 & \textbf{24} \\
& College
& 91 / 65 & 45 / 55 & 96 / 71 & 15 / 52 & \textbf{24} \\
& Graduate
& 98 / 68 & 46 / 55 & 94 / 68 & 24 / 50 & \textbf{23} \\
\addlinespace[3pt]
\multirow{5}{=}{Region}
& Capital area
& 88 / 69 & 50 / 53 & 98 / 70 & 14 / 49 & \textbf{21} \\
& Yeongnam
& 89 / 64 & 46 / 52 & 97 / 71 & 12 / 51 & \textbf{24} \\
& Honam
& 89 / 73 & 41 / 60 & 98 / 69 & 16 / 56 & \textbf{26} \\
& Chung\-cheong
& 88 / 70 & 41 / 57 & 96 / 73 & 5 / 53 & \textbf{26} \\
& Gangwon--Jeju
& 95 / 69 & 38 / 57 & 97 / 65 & 14 / 53 & \textbf{29} \\
\midrule
\multicolumn{2}{@{}l}{\textbf{Mean absolute gap}}
& \textbf{21} & \textbf{11} & \textbf{29} & \textbf{40} & \textbf{25} \\
\bottomrule
\end{tabularx}
% ---------- Panel B ----------
\begin{tabularx}{\textwidth}{@{}p{1.55cm}p{1.65cm}YYYYY@{}}
\multicolumn{7}{@{}l}{}\\%{\textbf{(b) Low-birthrate policy}} \\[2pt]
\toprule
\multicolumn{2}{l}{}
\textbf{(b) Low-birthrate policy}
& Work--family
& Educ.--care
& Econ.\ support
& Housing
& \textbf{Gap} \\
\midrule
\multirow{2}{=}{Sex}
& Male
& 99 / 53 & 73 / 58 & 19 / 57 & 98 / 50 & \textbf{37} \\
& Female
& 100 / 69 & 84 / 59 & 3 / 56 & 98 / 57 & \textbf{38} \\
\addlinespace[3pt]
\multirow{3}{=}{Age}
& 20s
& 100 / 62 & 62 / 61 & 5 / 62 & 100 / 46 & \textbf{38} \\
& 30s
& 99 / 60 & 89 / 62 & 13 / 58 & 100 / 53 & \textbf{40} \\
& 40s
& 99 / 62 & 79 / 55 & 10 / 53 & 100 / 57 & \textbf{37} \\
\addlinespace[3pt]
\multirow{2}{=}{Marital status}
& Single
& 100 / 62 & 60 / 62 & 10 / 61 & 100 / 49 & \textbf{35} \\
& Married
& 99 / 60 & 88 / 55 & 13 / 53 & 97 / 57 & \textbf{38} \\
\midrule
\multicolumn{2}{@{}l}{\textbf{Mean absolute gap}}
& \textbf{39} & \textbf{18} & \textbf{46} & \textbf{46} & \textbf{37} \\
\bottomrule
\end{tabularx}
\caption{
\textbf{Pre-deliberation persona responses vs. human survey benchmarks, by demographic group.}
Cells report the A-share as \textbf{Full persona / Human survey} (\%). Mean absolute gaps are reported by group (final column) and by issue (bottom row).}
\label{tab:bygroup}
\end{table*}

% =====================================================================
% 원설문 형식 복제 실험 (Appendix J).  Response instruments 절 뒤에 위치.
% 표 3개: J.1 = Table 4 형식, J.2 = Table 3 형식, J.3 = 셀 내 분산.
% 수치 출처: origformat-appendix/tables.json (원자료에서 재계산 검증 완료)
\section{Asking the personas the human questionnaire}
\label{app:origformat}
The two-position survey of Section~\ref{sec:instrument} is an instrument we designed for a language model, while the human respondents answered ranked-choice items with five to nine options. Here GPT-6-sol personas receive each source item with its original options, in the original order and in a random order drawn per persona (the scale item swaps its anchors), and reply with their first and second choices. Their answers are converted to an A-share with the function applied to the human respondents: the published recode for the environmental items, recomputed from the KEI respondent-level microdata ($n{=}3{,}008$),\footnote{The recomputation reproduces the published group shares in 63 of 64 cells to within rounding; the exception is graduate education on climate technology, 50.0\% in the microdata against 53 in the published table, and we use the recomputed value.} and the first-and-second-choice renormalization of the survey's own report for the low-birthrate items, for which no microdata exist. Each arm is 5{,}120 calls with no failures.

\paragraph{The original format does not close the gap.} On the seven items with a defined A-share under every instrument, the mean absolute gap from the population is 31 points in the original order and 33 shuffled, against 29 for the forced choice (Table~\ref{tab:origformat-instrument}). On housing, only 2.7\% of personas place either benchmark option in their top two, while 98.8\% place dedicated housing supply there, against 33.7\% of human respondents.

\paragraph{Personas in the same demographic cell answer alike.} Two human respondents from the same sex $\times$ age $\times$ education $\times$ region cell (seven survey regions for humans, five region groups for personas) select the identical pair of options 3.7 to 7.6\% of the time; two GPT-6-sol personas from the same cell do so 46 to 47\% of the time, and GPT-4.1-mini personas 51 to 93\% (Table~\ref{tab:origformat-dispersion}). GPT-6-sol personas also use about half as many distinct option pairs as humans, and GPT-4.1-mini personas about a quarter. The human spread comes from differences between individuals within groups, which the personas do not reproduce, so the failure is not an artifact of aggregating over groups.

\begin{table}[!htbp]
\centering\small\setlength{\tabcolsep}{4pt}
\resizebox{\columnwidth}{!}{\begin{tabular}{lcccc}\toprule
Question & Human & Forced choice & Original & Shuffled \\\midrule
Env.\ priority & 68 & 90 & 95 & 97 \\
Env.\ means & 54 & 43 & 71 & 76 \\
Clim.\ strategy & 70 & 97 & 50 & 42 \\
Clim.\ tech & 51 & 12 & 94 & 95 \\
Work--family & 61 & 100 & 89 & 92 \\
Educ.--care & 58 & 79 & 91 & 88 \\
Econ.\ support & 56 & 11 & 9 & 9 \\
Housing & 53 & 98 & 100$^\dagger$ & 100$^\dagger$ \\
\midrule
Mean abs.\ gap (7 items) & -- & 29 & 31 & 33 \\
\bottomrule\end{tabular}}
\caption{Original questionnaire against the natural-language forced choice, GPT-6-sol (A-share, \%). \emph{Original} asks the source item with its original options and applies the human scoring; \emph{shuffled} lists the options in a random order per persona. $^\dagger$Defined by the 17 (original) and 50 (shuffled) personas who place a benchmark option in their top two; excluded from the mean.}
\label{tab:origformat-instrument}
\end{table}

\begin{table}[!htbp]
\centering\small\setlength{\tabcolsep}{3pt}
\resizebox{\columnwidth}{!}{\begin{tabular}{l ccc ccc}\toprule
& \multicolumn{3}{c}{Same pair} & \multicolumn{3}{c}{Pairs} \\\cmidrule(lr){2-4}\cmidrule(lr){5-7}
Question & Human & GPT-6-sol & GPT-4.1-mini & Human & GPT-6-sol & GPT-4.1-mini \\\midrule
Env.\ means (9 options) & 3.7\% & 46.3\% & 50.8\% & 36 & 19 & 11 \\
Clim.\ strategy (8 options) & 5.3\% & 46.7\% & 92.9\% & 28 & 12 & 6 \\
Clim.\ tech (8 options) & 7.6\% & 46.9\% & 60.5\% & 25 & 14 & 5 \\
\bottomrule\end{tabular}}
\caption{Within-cell dispersion under the original questionnaire. \emph{Same pair}: probability that two respondents from the same sex $\times$ age $\times$ education $\times$ region cell (seven survey regions for humans, five region groups for personas) select the identical pair of options; \emph{pairs}: distinct option pairs the group ever selects.}
\label{tab:origformat-dispersion}
\end{table}

\section{Robustness across models and populations}
\label{app:robust}
The survey-stage failure holds across models, populations, and response instruments.

\paragraph{Six models on all eight issues.} Table~\ref{tab:xmodel} runs the natural-choice instrument on all six models, the full 640-persona pool, and all eight questions; these answers are the round-0 positions from which Section~\ref{sec:deliberation} builds its rooms. Every model sits 23 to 31 points from the population by question (21 to 30 by group), and the primary model, GPT-6-sol, is no closer (31) than GPT-4.1-mini (30). No model gets the direction of demographic differences right in a majority of the 110 comparisons (29 to 51).

\begin{table*}[!tbp]
  \centering\small
  \resizebox{\textwidth}{!}{\begin{tabular}{@{}lccccccc@{}}
    \toprule
    Question & Human & GPT-6-sol & GPT-6-luna & GPT-4.1-mini & DeepSeek-V4.1 & Claude-Haiku-5.5 & Qwen-3.8-flash \\
    \midrule
    Env.\ priority & 68 & 90 & 100 & 97 & 95 & 81 & 69 \\
    Env.\ means & 54 & 43 & 39 & 1 & 35 & 4 & 43 \\
    Clim.\ strategy & 70 & 97 & 95 & 91 & 84 & 81 & 100 \\
    Clim.\ tech & 51 & 12 & 68 & 67 & 46 & 71 & 48 \\
    Work--family & 61 & 100 & 91 & 100 & 80 & 85 & 100 \\
    Educ.--care & 58 & 79 & 83 & 61 & 79 & 43 & 92 \\
    Econ.\ support & 56 & 11 & 2 & 10 & 9 & 2 & 7 \\
    Housing & 53 & 98 & 100 & 23 & 21 & 39 & 16 \\
    \midrule
    Mean gap from human & --- & 31 & 31 & 30 & 23 & 25 & 25 \\
    Directions & --- & 49/51 & 34/64 & 29/69 & 47/61 & 51/56 & 47/44 \\
    \bottomrule
  \end{tabular}}
\caption{Cross-model natural-choice survey: A-share (\%) on each question for six models against the population (temperature 0, 640 personas). The GPT-6-sol column is the main survey of Section~\ref{sec:respondents}. \emph{Directions}: group-pair comparisons matched / reversed, of 110.}
\label{tab:xmodel}
\end{table*}

\paragraph{GPT-6-sol under the alternative setups.} Table~\ref{tab:solrobust} repeats the controls of Section~\ref{sec:survey-analysis} on GPT-6-sol under alternative response instruments: a neutral option, a labeled choice with counterbalanced order (both allow a neutral answer, which is excluded from the A-share), and two independent 0--2 ratings. The gap stays between 29 and 32 points under every setup except the two-rating format, which rises to 36 because on seven of eight questions more than 99\% of personas rate both opposing positions at 1 or higher. Given the human questionnaire (Appendix~\ref{app:origformat}), GPT-6-sol sits 31 points from the population on the seven comparable items (33 with the options shuffled, 29 under the forced choice), and on housing only 2.7\% of personas place either benchmark option in their top two.

\begin{table}[!htbp]
  \centering\small
  \resizebox{\columnwidth}{!}{%
\begin{tabular}{lccc}
    \toprule
    Setup & Gap, question & Gap, group & Directions \\
    \midrule
    Natural choice, temperature 0 (main) & 31 & 29 & 49/51 \\
    Natural choice, temperature 1 (5 samples) & 31 & 29 & 44/64 \\
    Demographics only (no narrative) & 30 & 28 & 44/56 \\
    Natural choice with a neutral option & 32 & 31 & 46/55 \\
    Labeled choice, counterbalanced & 29 & 27 & 54/54 \\
    Two independent ratings (0--2 each) & 36 & 38 & 28/33 \\
    \bottomrule
  \end{tabular}}
\caption{GPT-6-sol, 640 personas, eight questions: mean absolute gap of the A-share from the population (points), by question as in Table~\ref{tab:controls} and by demographic group as in Section~\ref{sec:validity}, and group-pair directions matched / reversed (of 110). The demographics-only, neutral-option, labeled, and two-rating setups are sampled at temperature 0.7.}
\label{tab:solrobust}
\end{table}

\paragraph{Reweighting to the survey's composition.} The persona pool is balanced, so a persona group such as men averages over a different mix of age, education, and region than the survey's men. Reweighting each GPT-6-sol persona cell to the survey's joint sex $\times$ age $\times$ education $\times$ region composition within each group, using the KEI microdata for the four environmental questions, barely changes the result: the mean absolute gap moves from 25.0 to 24.2 points over the 60 group--question cells, and the personas still reverse more group differences than they match (49 against 42 of 92, against 45 and 41 unweighted).

\paragraph{United States pool.} A parallel English pool of US personas benchmarked against Pew shows the same failure in a second language and population (Table~\ref{tab:us}): GPT-6-sol personas saturate at 97 to 100 percent on three of five items. The mean gap of 33 points is close to the Korean pool's 31 by question (GPT-4.1-mini: 28 against 30).

\begin{table}[!htbp]
  \centering
  \small
  \resizebox{\columnwidth}{!}{%
\begin{tabular}{lccc}
    \toprule
    Question & GPT-6-sol & GPT-4.1-mini & Human survey (Pew) \\
    \midrule
    Environmental regulation & 100 & 97  & 61 \\
    Energy priority           & 98  & 100 & 68 \\
    Abortion                  & 97  & 95  & 61 \\
    Size of government        & 80  & 45  & 49 \\
    Energy strategy           & 0   & 0   & 31 \\
    \midrule
    Mean abs.\ gap (pts)      & 33  & 28  & -- \\
    \bottomrule
  \end{tabular}}
\caption{U.S. persona responses against survey benchmarks. Responses are from 256 U.S. personas from Nemotron-Personas-USA, prompted in English and benchmarked against published Pew survey responses. Cells report the share choosing the first position (A-share, \%); gaps are computed from unrounded values.}
\label{tab:us}
\end{table}

\section{Prompts}
\label{app:prompts}
Agent prompts are in Korean; we give English translations. The personas come from Nemotron-Personas-Korea and Nemotron-Personas-USA (CC BY 4.0); the KEI microdata and the Shin-Kori first-wave data are publicly released for research use. Judge prompts are in English and given verbatim. Korean originals are in the released code. All models are called through OpenRouter (\texttt{openai/gpt-6-sol}, \texttt{openai/gpt-6-luna}, \texttt{openai/gpt-4.1-mini}, \texttt{deepseek/deepseek-v4.1-flash}, \texttt{anthropic/claude-haiku-5.5}, \texttt{qwen/qwen3.8-flash}; judges \texttt{anthropic/claude-opus-5.5}, \texttt{google/gemini-3.1-pro-preview}, and \texttt{openai/gpt-6-sol}) between August and October 2026, with reasoning disabled except for the Opus and Gemini judges, whose endpoints require it and run at minimal effort; unless noted, stance surveys use temperature 0 and 80 output tokens and speaking turns temperature 1 and 400.

\subsection{Shin-Kori panel}
The Shin-Kori runs use the templates of the policy debates below (survey, natural speaking turn, side cue, monologue) unchanged. Three things are filled from the real poll.
\paragraph{System prompt.} The profile line is filled from the respondent's first-wave record; records carry no narrative.
{\scriptsize\begin{verbatim}
You are a South Korean citizen with the profile below.
[Profile] Participant {k}, {age}-year-old {man/woman},
lives in {province}, occupation {occupation},
supports {party} / supports no party

Always answer only in the specified JSON format.
\end{verbatim}}
\paragraph{Issue and positions.}
{\scriptsize\begin{verbatim}
[Issue] Whether to build Shin-Kori units 5 and 6 (August 2017)
Position 1: Construction of Shin-Kori units 5 and 6
            should resume
Position 2: Construction of Shin-Kori units 5 and 6
            should be halted
\end{verbatim}}
\paragraph{Materials.} The two summaries, written for this study, state each side's main arguments at the same length (339 and 342 Korean characters). From the booklet stage on, every survey and speaking prompt carries both, in an order drawn at random per agent and kept for all of its prompts (249 of 498 agents see the resume side first), as a block after the issue line. Agents are named Participant 1--6 within a room, and each agent's profile line carries its own label, as Section~\ref{sec:deliberation} personas carry their name in the narrative.
{\scriptsize\begin{verbatim}
[Materials]
[Resume side] Construction is 28.8% complete; 1.6 trillion
won has been spent, and halting would cost 2.8 trillion won
including compensation. Total cost is 8.6 trillion won. The
APR1400 design reflects post-Fukushima standards with
magnitude-7.0 seismic design. Reducing nuclear power makes
the reserve margin unstable and weakens base-load supply;
solar and wind vary with the weather. Electricity prices
would rise for households and firms. Without domestic
construction, nuclear exports lose competitiveness. Ulju
Seosaeng residents voted to host the plant, and local
suppliers and jobs depend on it. It emits less greenhouse
gas and fine dust than coal.

[Halt side] As at Fukushima in 2011, a nuclear accident
cannot be undone. A magnitude-5.8 earthquake struck
Gyeongju in 2016, and the Yangsan fault may be active. About
3.8 million people live within 30 km of Kori, the densest
nuclear site in the world, and evacuation plans for Busan and
Ulsan are unrealistic. Spent-fuel storage is nearly full,
no disposal site has been chosen, and waste must be managed
for hundreds of thousands of years. Renewable costs keep
falling and countries such as Germany are phasing out
nuclear power. KHNP has a record of procurement corruption
and forged parts. Two more units lock the country into
nuclear power for 60 years. The long-term risk exceeds the
sunk cost.
\end{verbatim}}
\paragraph{Restated condition.} The side cue of the policy debates, in the round-1 speaking turn only, with the respondent's real wave-1 answer; respondents who answered undecided receive no cue.
{\scriptsize\begin{verbatim}
[Your position] You support the "{real wave-1 side}" side.
\end{verbatim}}
\paragraph{Three-way item.} At wave 1 and after round 3, agents also answer the same survey with a third option, ``I reserve judgment'', listed last; this answer is never shown to them.
\paragraph{Attribution test.} The booklet-stage survey with one statement in the discussion block, prefixed by ``You (your last turn):'', ``An expert who has studied this issue:'', ``Another participant in your group:'', or nothing.
\paragraph{Fictional names.} ``Shin-Kori units 5 and 6'' becomes ``Garam units 1 and 2''; the date is removed from the issue line, and the place names Ulju Seosaeng, Kori, and Busan and Ulsan in the summaries become ``a village near the site'', ``the Garam site'', and ``nearby large cities''. Fukushima, Gyeongju, the Yangsan fault, and the booklet's figures are unchanged.

\subsection{Policy debates}

\paragraph{Persona system prompt (all calls).}
{\scriptsize\begin{verbatim}
You are a South Korean citizen with the profile below.
[Profile] {age}-year-old {sex}, {region}, education {education},
occupation {occupation}, {household}.
[Background] {narrative}

Always answer only in the specified JSON format.
\end{verbatim}}

\paragraph{Attitude survey (private; before the debate and after each round).}
The bracketed transcript block is omitted before the debate and inserted after each round; the option order is randomized per persona and fixed across rounds.
{\scriptsize\begin{verbatim}
[Issue] {topic}
[Discussion so far]
{transcript}
Which of the two positions below are you closer to?
You must choose one.
- {position 1}
- {position 2}
Reply only in JSON, copying one position verbatim:
{"choice": "..."}
\end{verbatim}}

\paragraph{Deliberation speech turn, natural (nothing restated; Taubenfeld-faithful).}
Nothing from the agent's round-0 answer is surfaced. For the first speaker the transcript block reads ``No one has spoken yet.''
{\scriptsize\begin{verbatim}
[Issue] {topic}
[Positions] {position 1}  or  {position 2}
[Discussion so far]
{transcript}
You are {name}. Continue the discussion above and state your
view in 2-4 sentences. Reply with the utterance only, as JSON:
{"public": "..."}
\end{verbatim}}

\paragraph{Sealed monologue.} The natural speech turn and the attitude survey, with the discussion block holding only the agent's own earlier turns. \emph{Re-ask}: the attitude survey with no discussion block, asked three more times.

\paragraph{Deliberation speech turn, position restated.}
Identical to the natural speech turn, plus one line naming the agent's round-0 side, inserted between the positions and the transcript. This line is present in the round-1 prompt only; rounds 2--3 use the natural speech turn above.
{\scriptsize\begin{verbatim}
[Issue] {topic}
[Positions] {position 1}  or  {position 2}
[Your position] You support the "{round-0 side}" side.
[Discussion so far]
{transcript}
You are {name}. Continue the discussion above and state your
view in 2-4 sentences. Reply with the utterance only, as JSON:
{"public": "..."}
\end{verbatim}}

\paragraph{Argued opening.}
A dedicated round-0 turn, before round~1, in which the agent argues its assigned side; the resulting opening is placed in the transcript that round~1 reads.
{\scriptsize\begin{verbatim}
[Issue] {topic}
[Positions] {position 1}  or  {position 2}
[Your position] You are on the "{assigned side}" side.
You are {name}. State the reasons supporting this position in
2-4 sentences. Reply with the utterance only, as JSON:
{"public": "..."}
\end{verbatim}}

\paragraph{Other prompts.} The demographics-only persona replaces the profile and background with one line of sex, age band, education, and region. The Korean-citizen system prompt is ``You are a South Korean citizen. Always answer only in the specified JSON format.'', and the no-persona prompt keeps only its second sentence; both are asked 80 times per question at temperature 1, with the option order drawn per sample. The neutral-option survey adds ``If neither, `neutral'.''; the labeled survey lists the positions as (A) and (B) and also allows `neutral'; each rating prompt asks how much the persona supports one position, from 0 (not at all) to 2 (strongly). These three run at temperature 0.7, with 80, 30, and 25 output tokens. The U.S. survey is the attitude survey in English, without a discussion block. In the decomposition run (Appendix~\ref{app:anchor}), a free round-0 opening asks the agent which position it is closer to and why, in 2--4 sentences, and the survey begins ``This is a private survey separate from the discussion.'' The knowledge question, asked with no system prompt at temperature 1 (ten samples per model), is ``In the 2017 Shin-Kori 5 and 6 public deliberation, which did the citizen panel finally recommend, resuming or halting construction, and in what proportion? Answer in one line.''

\paragraph{Discourse quality judge.} System: ``You are an expert coder applying the Discourse Quality Index (DQI, Steenbergen et al. 2003) to a Korean deliberation transcript. Score ONLY the target utterance. Respond in JSON.''
{\scriptsize\begin{verbatim}
[Issue] {topic}
[Position 1] {position A}
[Position 2] {position B}
[Prior discussion]
{transcript before the turn, or (none)}
[TARGET utterance by {name}]
{utterance}

Score the TARGET only:
- justification: 0 no reason, 1 one incomplete reason,
  2 one complete reason with linkage, 3 sophisticated/multiple
- common_good: 0 self or group interest, 1 neutral,
  2 explicit common good / society
- respect: 0 dismissive of others, 1 neutral,
  2 explicit respect for others or counterarguments
- reciprocity: 0 ignores others, 1 engages a specific prior point
- novelty: "new" introduces a new argument, "build" elaborates
  a prior one, "repeat" restates
JSON: {"justification":0,"common_good":0,"respect":0,
       "reciprocity":0,"novelty":"new"}
\end{verbatim}}

\paragraph{Argument repertoire judge.} System: ``You extract distinct, non-redundant argument reasons from a participant's statements. Respond in JSON.''
{\scriptsize\begin{verbatim}
[Issue] {topic}
[Position A] {position A}
[Position B] {position B}
[Statements by {name}]
- {statement 1}
- ...
List the DISTINCT reasons this participant articulated
supporting Position A, and supporting Position B. Count a
reason once even if repeated; do not invent reasons not stated.
JSON: {"for_A":["..."],"for_B":["..."]}
\end{verbatim}}

\section{Starting composition and endpoints}
\label{app:composition}

\paragraph{Other models.} Table~\ref{tab:comp6} gives the same test for GPT-4.1-mini and GPT-6-luna on climate technology, education--care, and housing, with up to four unanimous rooms per side and the 20 balanced rooms of Section~\ref{sec:deliberation}. Their endpoints again depart from the population: on education--care, GPT-6-luna's rooms in which all six agents start on Position~B end with six on Position~A, above the population's 3.5, and on housing GPT-4.1-mini's all-B rooms end with 4.3, above the population's 3.2 and above its all-A rooms (3.8). In some cells the starting composition carries through, as it does for GPT-6-sol on climate technology: GPT-6-luna on climate technology and GPT-4.1-mini on education--care.
\begin{table}[!htbp]
\centering\footnotesize\setlength{\tabcolsep}{2.5pt}
\resizebox{\columnwidth}{!}{%
\begin{tabular}{ll ccc ccc}
\toprule
& & \multicolumn{3}{c}{Start} & & & \\
\cmidrule(lr){3-5}
Model & Question & All A & 3--3 & All B & Pop. & Pers. & Prior \\
\midrule
GPT-4.1-mini & Clim.\ tech & 2.8 & 2.6 & 1.5 & 3.0 & 4.0 & 5.9 \\
 & Educ.--care & 6.0 & 5.1 & 3.0 & 3.5 & 3.7 & 3.0 \\
 & Housing & 3.8 & 3.9 & 4.3 & 3.2 & 1.4 & 0.0 \\
\midrule
GPT-6-luna & Clim.\ tech & 6.0 & 4.2 & 2.3 & 3.0 & 4.1 & 6.0 \\
 & Educ.--care & 6.0 & 6.0 & 6.0 & 3.5 & 5.0 & 5.2 \\
 & Housing & 6.0 & -- & -- & 3.2 & 6.0 & 6.0 \\
\bottomrule
\end{tabular}}
\caption{Agents on Position~A (of six) after three rounds by starting composition, against the population benchmark, the personas' survey answers (Pers.), and the model's no-persona answer (Prior) (60 samples at temperature 1, both option orders), all scaled to six. GPT-6-luna has too few housing personas on Position~B for a balanced or all-B room.}
\label{tab:comp6}
\end{table}

\section{Anchoring and the source of movement}
\label{app:anchor}
\label{app:inject}
\begin{table*}[!tbp]
    \centering
  \small
  \setlength{\tabcolsep}{4pt}
  \begin{tabular}{l ccc ccc}
    \toprule
    & \multicolumn{3}{c}{Survey-chosen position} & \multicolumn{3}{c}{Population-matched position} \\
    \cmidrule(lr){2-4}\cmidrule(lr){5-7}
    Issue & Nothing & Position & Opening & Nothing & Position & Opening \\
          & restated & restated & argued & restated & restated & argued \\
    \midrule
    Env.\ priority (saturated) & 6.0 (50\%) & 3.0 (0\%) & 3.0 (0\%) & 6.0 (50\%) & 3.0 (0\%) & 3.0 (0\%) \\
    Env.\ means (divisive) & 1.7 (50\%) & 3.0 (0\%) & 3.0 (0\%) & 2.3 (56\%) & 3.0 (0\%) & 3.0 (0\%) \\
    Clim.\ strategy (saturated) & 6.0 (50\%) & 3.0 (0\%) & 3.0 (0\%) & 6.0 (50\%) & 3.5 (8\%) & 3.0 (0\%) \\
    Clim.\ tech (saturated) & 1.2 (42\%) & 3.0 (0\%) & 3.0 (0\%) & 1.0 (50\%) & 3.0 (0\%) & 3.0 (0\%) \\
    Work--family (saturated) & -- & -- & -- & 6.0 (50\%) & 3.0 (0\%) & 3.0 (0\%) \\
    Educ.--care (saturated) & 5.7 (44\%) & 3.0 (0\%) & 3.0 (0\%) & 5.8 (47\%) & 3.0 (0\%) & 3.0 (0\%) \\
    Econ.\ support (saturated) & 0.0 (50\%) & 3.0 (0\%) & 3.0 (0\%) & 0.0 (50\%) & 3.0 (0\%) & 3.0 (0\%) \\
    Housing (saturated) & 4.8 (54\%) & 3.0 (0\%) & 3.0 (0\%) & 6.0 (50\%) & 3.0 (0\%) & 3.0 (0\%) \\
    \bottomrule
  \end{tabular}
\caption{Final room composition under six protocols varying position assignment and anchoring. An agent's position is either its own survey answer (\emph{survey-chosen}) or assigned using the population benchmarks while keeping the room balanced (\emph{population-matched}), and before the debate it is given nothing, a one-line assignment of its position, or a self-argued opening. Cells report the mean number of agents supporting Position~A after three rounds, with the percentage that changed position. Dashes: fewer than three personas chose Position~B on work--family, so no balanced survey-chosen room exists.}
\label{tab:2x3}
\end{table*}

Table~\ref{tab:2x3} varies how each agent's starting position is assigned and whether it appears in the debate (GPT-6-sol, up to six rooms per question and cell). An agent's position is either its own survey answer (\emph{survey-chosen}) or assigned from the population benchmarks with the room kept balanced (\emph{population-matched}: each persona's side is drawn with a probability built from its demographic groups' survey shares for Position~A, the overall share shifted on the logit scale by each group's difference from it and clipped to 5--95\%, and rooms are then filled three per side). Without the position restated, agents change side often under either assignment. When it is restated in one line or argued in an opening, movement falls to 0\% in all but one cell. A separate run decomposes the hold (GPT-6-sol; education--care, housing, and environmental priority, five rooms each, in which the three personas most likely by the benchmarks to favor Position~A are assigned to it; every arm opens with a round-0 turn, free or arguing the assigned side; speaking turns at temperature 0.7, and the survey states that it is private and separate from the discussion). With the assigned side argued in the opening, no agent leaves it; hiding one same-side peer's turns from the survey still leaves 0\% moving, while hiding the agent's own turns restores movement to 47\%. When the side is assigned but never shown to the agent, 56\% end on the other side, as in the arm with a free opening and no assignment (56\%).

\paragraph{Six models.} Table~\ref{tab:grid6} runs the same grid on all six models, up to six rooms per question and cell, and asks every survey a second time with the agent's own turns removed from the transcript.
\begin{table*}[!tbp]
\centering\small\setlength{\tabcolsep}{4pt}
\resizebox{\textwidth}{!}{\begin{tabular}{lc ccc ccc cccc}
\toprule
& & \multicolumn{3}{c}{Survey-chosen} & \multicolumn{3}{c}{Population-matched} & \multicolumn{4}{c}{Own turns removed} \\
\cmidrule(lr){3-5}\cmidrule(lr){6-8}\cmidrule(lr){9-12}
Model & Rooms & Nothing & Position & Opening & Nothing & Position & Opening & \multicolumn{2}{c}{Chosen: Pos./Open.} & \multicolumn{2}{c}{Matched: Pos./Open.} \\
\midrule
GPT-6-sol & 261 & 48 & 0 & 0 & 50 & 1 & 0 & 14 & 16 & 22 & 26 \\
GPT-6-luna & 246 & 49 & 15 & 12 & 49 & 22 & 20 & 65 & 61 & 55 & 59 \\
GPT-4.1-mini & 255 & 40 & 1 & 1 & 49 & 1 & 3 & 47 & 50 & 58 & 60 \\
DeepSeek-V4.1 & 288 & 44 & 9 & 3 & 50 & 15 & 9 & 47 & 52 & 58 & 59 \\
Claude-Haiku-5.5 & 285 & 46 & 11 & 13 & 49 & 27 & 22 & 43 & 49 & 48 & 54 \\
Qwen-3.8-flash & 252 & 44 & 5 & 8 & 48 & 8 & 9 & 32 & 37 & 42 & 48 \\
\bottomrule
\end{tabular}}
\caption{Agents changing side over three rounds (\%), by how the starting position is assigned and carried into the debate. \emph{Own turns removed}: the same rooms, surveyed without the agent's own turns in the transcript.}
\label{tab:grid6}
\end{table*}

\section{Discourse quality across judges}
\label{app:dqi}
Three frontier models score the same 35 GPT-6-sol debates (five per question; 630 turns) with the Discourse Quality Index \citep{steenbergen2003} and argument repertoire \citep{cappella2002}, at temperature 0: Claude-Opus-5.5 and Gemini-3.1-pro, from families other than the debater's (with minimal reasoning, which their endpoints require), and GPT-6-sol itself. Each turn is coded against the discussion before it on three DQI indicators, level of justification (0--3), content of justification (0--2: group interest, neutral, common good), and respect (0--2, merging the DQI's respect toward groups, demands, and counterarguments), and on reciprocity (0--1: whether the turn responds to a specific earlier point; \citealp{graham2003,stromergalley2007}). The judge prompts are in Appendix~\ref{app:prompts}. GPT-6-sol rates the justification of its own debates at 2.75 of 3, against 1.95 from the two independent judges, so the main text reports the mean of Claude-Opus-5.5 and Gemini-3.1-pro (Table~\ref{tab:dqijudges}). The two independent judges agree on respect and reciprocity (weighted $\kappa$ 0.66 and 0.71; 0.27 on justification and 0.43 on common good). Agreement with GPT-6-sol is lower except for Gemini-3.1-pro on common good: $\kappa$ is 0.11, 0.39, 0.42, and 0.45 for Claude-Opus-5.5 and 0.08, 0.63, 0.58, and 0.56 for Gemini-3.1-pro on justification, common good, respect, and reciprocity.

The same two judges score the sealed-monologue twins of these rooms, where a turn is judged against the agent's own earlier turns, which is all it saw. The monologues score at least as high as the debates on justification, common good, and respect, under each judge. Averaged over rounds 1--3, the mean of the two judges is 1.95 for the debates against 2.15 for the monologues on justification, 1.34 against 1.43 on common good, and 1.25 against 1.28 on respect (Claude-Opus-5.5: 1.80/2.00, 1.22/1.30, 1.21/1.25; Gemini-3.1-pro: 2.10/2.31, 1.47/1.56, 1.29/1.32). Reciprocity is not defined for a monologue.

\paragraph{Human annotation.} Two of the authors, both native Korean speakers, independently scored 105 of the 630 debate turns, one drawn at random from each round of each of the 35 rooms, with the judges' rubric and a short written clarification of each scale, without seeing the judges' scores for these turns. Weighted $\kappa$ between the two annotators is 0.19 on justification, 0.24 on common good, 0.64 on respect, and 0.28 on reciprocity, and between either annotator and either judge it is 0.03 to 0.32, 0.15 to 0.23, 0.25 to 0.70, and 0.27 to 0.59; between the two judges on these turns it is 0.27, 0.49, 0.72, and 0.80. The judges' mean justification (1.83 and 2.14) lies between the annotators' (1.36 and 2.39), and both annotators score common good lower than both judges (0.93 and 0.59, against 1.21 and 1.48). Both annotators score justification lower in round 3 than in round 1 (from 1.97 to 1.11 and from 2.66 to 2.14) and reciprocity lowest in round 1.

\begin{table*}[!tbp]\centering\small
\resizebox{\textwidth}{!}{\begin{tabular}{l c ccc c}\toprule
& & \multicolumn{3}{c}{Each judge} & \\\cmidrule(lr){3-5}
Measure, debate turns & Mean of Opus and Gemini & Claude-Opus-5.5 & Gemini-3.1-pro & GPT-6-sol & $\kappa_w$, Opus--Gemini \\\midrule
Justification (0--3) & 2.10 / 1.82 & 1.99 / 1.63 & 2.22 / 2.01 & 2.84 / 2.69 & 0.27 \\
Common good (0--2) & 1.37 / 1.32 & 1.24 / 1.21 & 1.50 / 1.43 & 1.42 / 1.48 & 0.43 \\
Respect (0--2) & 1.25 / 1.20 & 1.21 / 1.18 & 1.29 / 1.23 & 1.43 / 1.50 & 0.66 \\
Reciprocity (0--1) & 0.70 / 0.94 & 0.69 / 0.91 & 0.71 / 0.96 & 0.83 / 1.00 & 0.71 \\
\midrule
Distinct reasons, first / final & 2.3 / 4.1 & 2.4 / 4.8 & 2.1 / 3.4 & 2.3 / 3.5 & \\
Reasons for both sides, final & 76\% & 77\% & 74\% & 70\% & \\
\bottomrule\end{tabular}}
\caption{Discourse quality of the same 630 debate turns (35 GPT-6-sol rooms) by three judges, round 1 / round 3. Reasons: distinct reasons per agent in its first turn and over all its turns.}
\label{tab:dqijudges}\end{table*}

\section{Discourse quality of live and frozen rooms}
\label{app:frozen-dqi}
Rooms in which no agent changes its mind score as high on discourse quality as a live room, or higher. We score ten rooms from each of four cells of Table~\ref{tab:2x3} with the two judges of Appendix~\ref{app:dqi}: the live natural room (survey-chosen side, nothing restated) and three frozen protocols, the argued opening (population-matched side) and the one-line restatement on the survey-chosen and on the population-matched pool. Every frozen room scores at least as high as the live room on justification, respect, and reciprocity, gives more distinct reasons in its first turn, and argues both sides more often; only common good is higher in the live room (Table~\ref{tab:frozen-dqi}). The transcript measures therefore do not reveal whether agents updated.

\begin{table}[!htbp]
  \centering\footnotesize
  \resizebox{\columnwidth}{!}{%
\begin{tabular}{lccccccc}
    \toprule
     & Moved & Justif.\ & Common & Respect & Recip.\ & Reasons & Both sides \\
    Room & (R0$\to$R3) & (0--3) & (0--2) & (0--2) & (0--1) & (first$\to$final) & (final) \\
    \midrule
    Natural, live & 47\% & 1.94 & 1.31 & 1.25 & 0.85 & 2.2$\to$4.1 & 61\% \\
    Argued opening & 0\% & 1.96 & 1.18 & 1.98 & 1.00 & 2.5$\to$4.4 & 87\% \\
    Position restated, survey-chosen & 0\% & 1.99 & 1.27 & 1.93 & 0.94 & 2.7$\to$4.0 & 85\% \\
    Position restated, population-matched & 0\% & 1.99 & 1.27 & 1.93 & 0.94 & 2.5$\to$4.0 & 87\% \\
    \bottomrule
  \end{tabular}}
\caption{Discourse quality and argument repertoire for a live room and three frozen protocols, ten GPT-6-sol rooms each: mean of Claude-Opus-5.5 and Gemini-3.1-pro, DQI averaged over rounds 1--3. \emph{Moved}: agents changing side between round 0 and round 3; \emph{reasons} and \emph{both sides}: distinct reasons per agent and share giving reasons for both sides, first turn to final.}
\label{tab:frozen-dqi}
\end{table}

\section{Debate versus monologue}
\label{app:graded}
\begin{table}[!htbp]
  \centering
  \small
  \begin{tabular}{lccc}
    \toprule
    Issue & Debate & Monologue & Difference \\
    \midrule
    Env.\ priority   & 5.70 & 5.70 & $0.00$ \\
    Env.\ means      & 2.50 & 1.95 & $+0.55$ \\
    Clim.\ strategy  & 5.95 & 6.00 & $-0.05$ \\
    Clim.\ tech      & 1.50 & 1.45 & $+0.05$ \\
    Educ.--care      & 5.55 & 5.80 & $-0.25$ \\
    Econ.\ support   & 0.05 & 0.05 & $0.00$ \\
    Housing          & 5.70 & 5.55 & $+0.15$ \\
    \bottomrule
  \end{tabular}
\caption{Debate versus sealed monologue: mean number of agents supporting Position A after three rounds, out of six agents, in balanced rooms, GPT-6-sol.}
\label{tab:mono}
\end{table}

Table~\ref{tab:mono} gives the GPT-6-sol rooms by question. Across the six models, agents change side in 41--50\% of debate seats and 39--44\% of monologue seats. Pooled debate$-$monologue differences in final count on Position~A have 90\% intervals of [$-$0.10, +0.23] (GPT-6-sol), [$-$0.20, +0.29] (GPT-6-luna), [$-$0.17, +0.29] (GPT-4.1-mini), [$-$0.03, +0.46] (DeepSeek), and [$-$0.10, +0.39] (Claude-Haiku), all within $\pm$0.5 agents; Qwen-3.8-flash's is [+0.16, +0.68]. Dropping questions with fewer than 30 distinct personas on the minority side, which rooms on saturated questions reuse, gives pooled differences of +0.07, +0.05, $-$0.13, +0.22, +0.34, and +0.42.

Room by room, the two conditions differ more than two runs of the same monologue. Table~\ref{tab:excess} subtracts, for each room, the absolute monologue-versus-rerun difference in final count from the absolute debate-versus-monologue difference; pooled over questions the excess is 0.2--0.6 agents, with a 95\% interval above zero in five of six models, and the largest per-question excess is on environmental means, the one divisive question for GPT-6-sol, in four of six models. Individual agents' final answers differ by 4.6 more points between debate and monologue than between two monologues for GPT-6-sol (95\% interval [2.1, 7.3]).
\begin{table*}[!tbp]\centering\small\setlength{\tabcolsep}{3.5pt}
\resizebox{\textwidth}{!}{\begin{tabular}{lccccccccc}\toprule
 & Env.\ priority & Env.\ means & Clim.\ strategy & Clim.\ tech & Work--family & Educ.--care & Econ.\ support & Housing & Pooled \\\midrule
GPT-6-sol & $-$0.05 & \textbf{+1.20} & +0.00 & +0.45 & -- & +0.25 & +0.10 & +0.30 & \textbf{+0.32} \\
GPT-6-luna & -- & \textbf{+1.30} & +0.00 & \textbf{+1.25} & +0.00 & +0.25 & $-$0.05 & -- & \textbf{+0.46} \\
GPT-4.1-mini & $-$0.10 & \textbf{+0.90} & $-$0.10 & +0.45 & -- & +0.35 & +0.05 & \textbf{+0.80} & \textbf{+0.34} \\
DeepSeek-V4.1 & \textbf{+0.70} & \textbf{+1.40} & +0.35 & $-$0.10 & +0.35 & +0.20 & \textbf{+0.60} & \textbf{+0.90} & \textbf{+0.55} \\
Claude-Haiku-5.5 & \textbf{+0.45} & \textbf{+1.00} & +0.00 & \textbf{+1.55} & +0.05 & +0.40 & +0.05 & \textbf{+1.25} & \textbf{+0.59} \\
Qwen-3.8-flash & +0.40 & $-$0.05 & -- & +0.45 & -- & +0.10 & $-$0.40 & +0.55 & +0.18 \\
\bottomrule\end{tabular}}
\caption{Excess of the absolute debate-versus-monologue difference over the absolute monologue-versus-rerun difference in final count on Position~A (agents of six), per question and pooled. Bold: 95\% interval from resampling rooms excludes zero. A dash marks a question with no balanced room.}
\label{tab:excess}\end{table*}

\section{Shin-Kori results with intervals}
\label{app:poll}
Table~\ref{tab:pollfull} gives each Shin-Kori run with the difference from the real panel and its 95\% interval (resampling rooms and redrawing the real wave, $n{=}471$), and the share of agents whose final answer differs from their booklet-stage answer. Unparsable answers are under 0.4\% in every run. Each condition reruns the pre-discussion surveys, so its wave-1 and booklet-stage shares differ slightly from the debate's; repeated temperature-0 calls change 5--30\% of answers (Section~\ref{sec:delib-surface}).
\begin{table*}[!tbp]
\centering\footnotesize
\setlength{\tabcolsep}{4pt}
\resizebox{\textwidth}{!}{\begin{tabular}{ll ccc c}
\toprule
& & Wave 1 & Booklet (wave 3) & After three rounds (wave 4) & Changed \\
\midrule
\multicolumn{2}{l}{\textbf{Real panel}} & 57.0 & 59.3 & 59.2 & -- \\
\midrule
GPT-6-sol & Debate & 47.0 ($-10.0$ [$-15.4$, $-4.3$]) & 41.0 ($-18.3$ [$-24.1$, $-12.5$]) & 44.2 ($-15.0$ [$-20.9$, $-9.1$]) & 14 \\
 & Monologue & 46.0 ($-11.0$ [$-16.9$, $-4.9$]) & 41.8 ($-17.5$ [$-23.8$, $-11.3$]) & 45.8 ($-13.4$ [$-19.3$, $-7.3$]) & 12 \\
 & Restated & 48.2 ($-8.8$ [$-14.3$, $-3.0$]) & 42.0 ($-17.3$ [$-23.3$, $-11.4$]) & 51.4 ($-7.8$ [$-13.4$, $-1.5$]) & 28 \\
 & Fictional names & 42.8 ($-14.2$ [$-20.2$, $-8.4$]) & 34.5 ($-24.7$ [$-30.6$, $-18.7$]) & 37.3 ($-21.9$ [$-27.6$, $-16.1$]) & 11 \\
\midrule
GPT-6-luna & Debate & 69.5 ($+12.5$ [$+6.4$, $+18.7$]) & 63.3 ($+4.0$ [$-1.8$, $+9.8$]) & 45.0 ($-14.2$ [$-21.5$, $-7.3$]) & 34 \\
 & Monologue & 73.9 ($+16.9$ [$+10.5$, $+23.0$]) & 68.1 ($+8.8$ [$+2.8$, $+15.0$]) & 47.6 ($-11.6$ [$-17.7$, $-5.2$]) & 35 \\
 & Restated & 73.1 ($+16.1$ [$+10.6$, $+21.6$]) & 63.1 ($+3.8$ [$-2.3$, $+9.7$]) & 52.8 ($-6.4$ [$-12.5$, $+0.1$]) & 34 \\
 & Fictional names & 43.2 ($-13.8$ [$-19.7$, $-7.8$]) & 29.3 ($-30.0$ [$-35.6$, $-23.9$]) & 18.7 ($-40.5$ [$-45.9$, $-35.1$]) & 23 \\
\midrule
GPT-4.1-mini & Debate & 43.2 ($-13.8$ [$-19.9$, $-7.2$]) & 38.6 ($-20.7$ [$-26.5$, $-14.9$]) & 49.1 ($-10.1$ [$-15.4$, $-4.9$]) & 26 \\
 & Monologue & 44.2 ($-12.8$ [$-18.8$, $-6.8$]) & 38.4 ($-20.9$ [$-27.1$, $-15.0$]) & 47.8 ($-11.4$ [$-17.5$, $-5.3$]) & 19 \\
 & Restated & 44.0 ($-13.0$ [$-19.3$, $-6.9$]) & 37.8 ($-21.5$ [$-27.2$, $-15.6$]) & 53.8 ($-5.4$ [$-11.1$, $+0.4$]) & 40 \\
 & Fictional names & 27.7 ($-29.3$ [$-34.8$, $-23.5$]) & 27.1 ($-32.2$ [$-37.6$, $-26.8$]) & 51.0 ($-8.2$ [$-13.9$, $-2.7$]) & 28 \\
\midrule
DeepSeek-V4.1 & Debate & 28.7 ($-28.3$ [$-34.0$, $-22.7$]) & 34.0 ($-25.3$ [$-31.0$, $-19.3$]) & 30.6 ($-28.6$ [$-35.6$, $-21.9$]) & 28 \\
 & Monologue & 30.5 ($-26.5$ [$-32.2$, $-20.8$]) & 31.1 ($-28.2$ [$-34.1$, $-22.4$]) & 27.9 ($-31.3$ [$-36.7$, $-25.5$]) & 18 \\
 & Restated & 29.6 ($-27.4$ [$-33.1$, $-21.9$]) & 31.5 ($-27.8$ [$-33.4$, $-22.3$]) & 43.2 ($-16.0$ [$-22.4$, $-9.8$]) & 32 \\
 & Fictional names & 14.5 ($-42.6$ [$-47.6$, $-37.0$]) & 28.9 ($-30.4$ [$-35.8$, $-24.9$]) & 30.9 ($-28.3$ [$-35.1$, $-21.0$]) & 22 \\
\midrule
Claude-Haiku-5.5 & Debate & 51.4 ($-5.6$ [$-11.6$, $+0.3$]) & 34.7 ($-24.5$ [$-30.2$, $-18.9$]) & 36.9 ($-22.3$ [$-30.2$, $-13.9$]) & 32 \\
 & Monologue & 51.2 ($-5.8$ [$-11.6$, $-0.2$]) & 35.5 ($-23.7$ [$-29.9$, $-17.8$]) & 45.6 ($-13.6$ [$-19.5$, $-7.7$]) & 31 \\
 & Restated & 51.6 ($-5.4$ [$-11.2$, $+0.2$]) & 35.1 ($-24.1$ [$-30.1$, $-18.5$]) & 51.8 ($-7.4$ [$-14.3$, $-0.7$]) & 40 \\
 & Fictional names & 34.1 ($-22.9$ [$-29.2$, $-16.8$]) & 16.7 ($-42.6$ [$-48.3$, $-37.1$]) & 20.3 ($-38.9$ [$-45.3$, $-32.6$]) & 19 \\
\midrule
Qwen-3.8-flash & Debate & 89.0 ($+31.9$ [$+26.6$, $+37.1$]) & 61.4 ($+2.2$ [$-3.8$, $+8.1$]) & 43.0 ($-16.2$ [$-22.2$, $-10.1$]) & 35 \\
 & Monologue & 89.4 ($+32.3$ [$+27.1$, $+37.4$]) & 60.2 ($+1.0$ [$-4.8$, $+7.0$]) & 48.6 ($-10.6$ [$-16.7$, $-4.5$]) & 27 \\
 & Restated & 88.0 ($+30.9$ [$+25.4$, $+36.5$]) & 61.8 ($+2.6$ [$-3.3$, $+8.8$]) & 52.8 ($-6.4$ [$-12.9$, $-0.4$]) & 31 \\
 & Fictional names & 90.4 ($+33.4$ [$+28.2$, $+38.5$]) & 65.5 ($+6.2$ [$+0.4$, $+12.0$]) & 43.2 ($-16.0$ [$-22.3$, $-10.0$]) & 34 \\
\bottomrule
\end{tabular}}
\caption{Resume share among agents choosing a side (\%), with the difference from the real wave and its 95\% interval. \emph{Changed}: final answer differs from the booklet-stage answer (\%).}
\label{tab:pollfull}
\end{table*}

\paragraph{Survey stage by group and by person.} Where two groups of real respondents (by sex, age, region, or party) differ by more than 5 points in their wave-1 answers, agents order them correctly in 21 (GPT-6-sol), 21 (GPT-6-luna), 18 (GPT-4.1-mini), 17 (DeepSeek), 18 (Claude-Haiku), and 17 (Qwen) of 21 comparisons, while a group's agents sit 10, 14, 16, 26, 10, and 32 points from its real members on average.

\paragraph{Three-way item.} With an undecided option, agents leave 11, 67, 37, 20, 46, and 31\% undecided at wave 1 (GPT-6-sol, GPT-6-luna, GPT-4.1-mini, DeepSeek, Claude-Haiku, Qwen), against 35.8\% of the panel, and 0, 0, 9, 2, 27, and 7\% after the debate, against 3.3\%. Total variation distance from the real three-way split after the debate is 0.13 to 0.29 (0.13 to 0.33 in the monologue), against 0.03 between two samples of the same distribution.

\paragraph{Restated condition.} The cue restates the real respondent's wave-1 answer, not the agent's own booklet-stage answer; the two differ for 30--49\% of cued agents, and 88--100\% of those end on the cued side. Respondents who were undecided at wave 1 receive no cue.

\paragraph{Argument attribution.} Share of agents moving to the side argued by the same statement, labeled as the agent's own last turn / unlabeled / an expert who has studied this issue / another participant in the group: GPT-6-sol 92.6 / 74.5 / 33.5 / 18.1\%, GPT-6-luna 91.8 / 91.0 / 76.3 / 57.0\%, GPT-4.1-mini 90.6 / 43.0 / 41.6 / 19.5\%, DeepSeek 82.7 / 37.3 / 21.5 / 9.8\%, Claude-Haiku 80.7 / 36.1 / 14.7 / 18.7\%, Qwen 84.9 / 42.4 / 16.3 / 11.2\% (498 agents each). Statements are round-1 statements, longer than 30 characters, of a separate GPT-6-sol debate run whose speakers chose the side they argue after round 1 and that do not mention another participant; the same statement goes to an agent under all four labels. Each agent's booklet-stage side comes from a separate run of its own model; agents without a two-sided booklet-stage answer receive a randomly chosen side, and in this test the profile line carries no participant label.

\end{document}